\documentclass[10pt,journal,compsoc]{IEEEtran}
\pdfoutput=1

\ifCLASSOPTIONcompsoc
  \usepackage[nocompress]{cite}
\else
  \usepackage{cite}
\fi

 \usepackage[pdftex]{graphicx}
\usepackage{color, soul, framed}
\usepackage{colortbl} 
\usepackage[dvipsnames, svgnames, x11names]{xcolor}
\usepackage{tabularx}
\usepackage{color,xcolor}
\usepackage{ amssymb }
\usepackage{amsmath,bm}
\usepackage{multirow} 
\usepackage{booktabs}
\usepackage{array}
\usepackage{algorithm}
\usepackage[colorlinks, linkcolor=red, anchorcolor=green, citecolor=blue]{hyperref}
 \usepackage[misc]{ifsym}
\usepackage{makecell}

\ifCLASSINFOpdf
 
\else
 
\fi

\begin{document}
\title{Deep Learning Methods for Calibrated Photometric Stereo and Beyond}

 \author{Yakun Ju, Kin-Man Lam, Wuyuan Xie, Huiyu Zhou, Junyu Dong,  and Boxin Shi  
        
\IEEEcompsocitemizethanks{\IEEEcompsocthanksitem Yakun Ju is with the Rapid-Rich Object Search Lab (ROSElab), School of Electrical and Electronic Engineering, Nanyang Technological University, Singapore (e-mail: kelvin.yakun.ju@gmail.com).

\IEEEcompsocthanksitem Kin-Man Lam is with the Department of Electrical and Electronic Engineering, The Hong Kong Polytechnic University, Hong Kong (e-mail: enkmlam@polyu.edu.hk).

\IEEEcompsocthanksitem Wuyuan Xie is with the Research Institute for Future Media Computing, Shenzhen University, Shenzhen, China (e-mail: wuyuan.xie@gmail.com).

\IEEEcompsocthanksitem Huiyu Zhou is with the Department of Informatics, University of Leicester, Leicester, UK (e-mail: hz143@leicester.ac.uk).

\IEEEcompsocthanksitem Junyu Dong is with the Faculty of Information Science and Engineering and the Institute for Advanced Ocean Study, Ocean University of China, Qingdao (e-mail: dongjunyu@ouc.edu.cn).

\IEEEcompsocthanksitem Boxin Shi is with the National Key Laboratory for Multimedia Information Processing and National Engineering Research Center
of Visual Technology, School of Computer Science, Peking University, Beijing, China (e-mail: shiboxin@pku.edu.cn).
}
}


\IEEEtitleabstractindextext{%
\begin{abstract}
Photometric stereo recovers the surface normals of an object from multiple images with varying shading cues, \emph{i.e.}, modeling the relationship between surface orientation and intensity at each pixel. Photometric stereo prevails in superior per-pixel resolution and fine reconstruction details. However, it is a complicated problem because of the non-linear relationship caused by non-Lambertian surface reflectance. Recently, various deep learning methods have shown a powerful ability in the context of photometric stereo against non-Lambertian surfaces. This paper provides a comprehensive review of existing deep learning-based calibrated photometric stereo methods utilizing orthographic cameras and directional light sources. We first analyze these methods from different perspectives, including input processing, supervision, and network architecture. We summarize the performance of deep learning photometric stereo models on the most widely-used benchmark data set. This demonstrates the advanced performance of deep learning-based photometric stereo methods. Finally, we give suggestions and propose future research trends based on the limitations of existing models.
\end{abstract}

}

\maketitle

\IEEEdisplaynontitleabstractindextext

\IEEEpeerreviewmaketitle

\IEEEraisesectionheading{\section{Introduction}\label{sec:introduction}}

\begin{figure}[t]
 \begin{center} \includegraphics[width=0.48\textwidth]{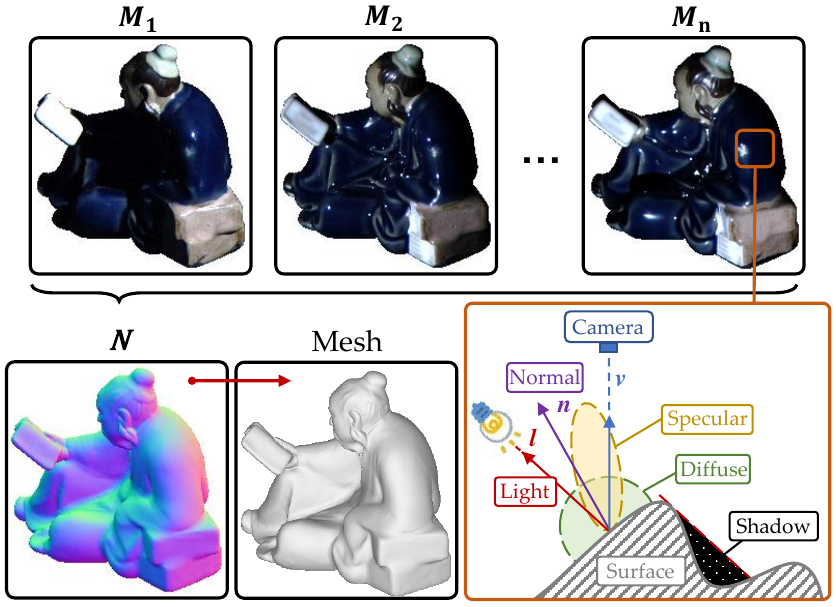}
  \end{center}
    \vspace{-3mm} 
  \caption{The schematic of photometric stereo. The orange box shows the general surface reflectance. }
  \label{figshowps}
  \vspace{-4mm} 
\end{figure}

\IEEEPARstart{A}{cquiring} three-dimensional (3D) geometry from two-dimensional (2D) scenes is a fundamental problem in computer vision. It aims to establish computational models that allow computers to perceive the external 3D world. Unlike geometric approaches (such as multi-view stereo and binocular) that use different viewpoint scenes to compute 3D points, photometric stereo \cite{Woodham1980Photometric} perceives the shape of an object from varying shading cues observed under different lighting conditions with a fixed viewpoint. Compared to geometric methods that generally reconstruct rough shapes, photometric methods can acquire more detailed local reconstruction. Therefore, photometric stereo plays a mainstream role in many high-precision surface reconstruction tasks, such as cultural relic reconstruction \cite{zhou2013multi}, seabed mapping \cite{fan2017refractive}, moon surface reconstruction \cite{wu2021centimeter}, and industrial defect detection \cite{ren2018fast}, \emph{etc}. As shown in Fig. \ref{figshowps}, photometric stereo methods obtain detailed shape reconstructions from multiple images under different illuminations. In this survey, we take the object ``Reading'' from the DiLiGenT benchmark \cite{shi2019benchmark} as a visual example, which has spatially varying and non-Lambertian materials with strong specularity and shadow.

Classic photometric stereo \cite{Woodham1980Photometric} assumed that only the Lambertian (diffuse) reflectance exists on the surface of the target object. Under the Lambertian assumption, the surface normal can be easily solved by the least squares method, because the reflection intensity $\boldsymbol{M}$ is linearly proportional to the angle between the normal $\boldsymbol{n}$  and incident light $\boldsymbol{l}$, as follows:
\begin{equation}
\label{proportion}
\boldsymbol{M} \propto \boldsymbol{l}^{\top} \boldsymbol{n}.
\end{equation}

\begin{figure*}[ht]
 \begin{center}
  \includegraphics[width=0.98\textwidth]{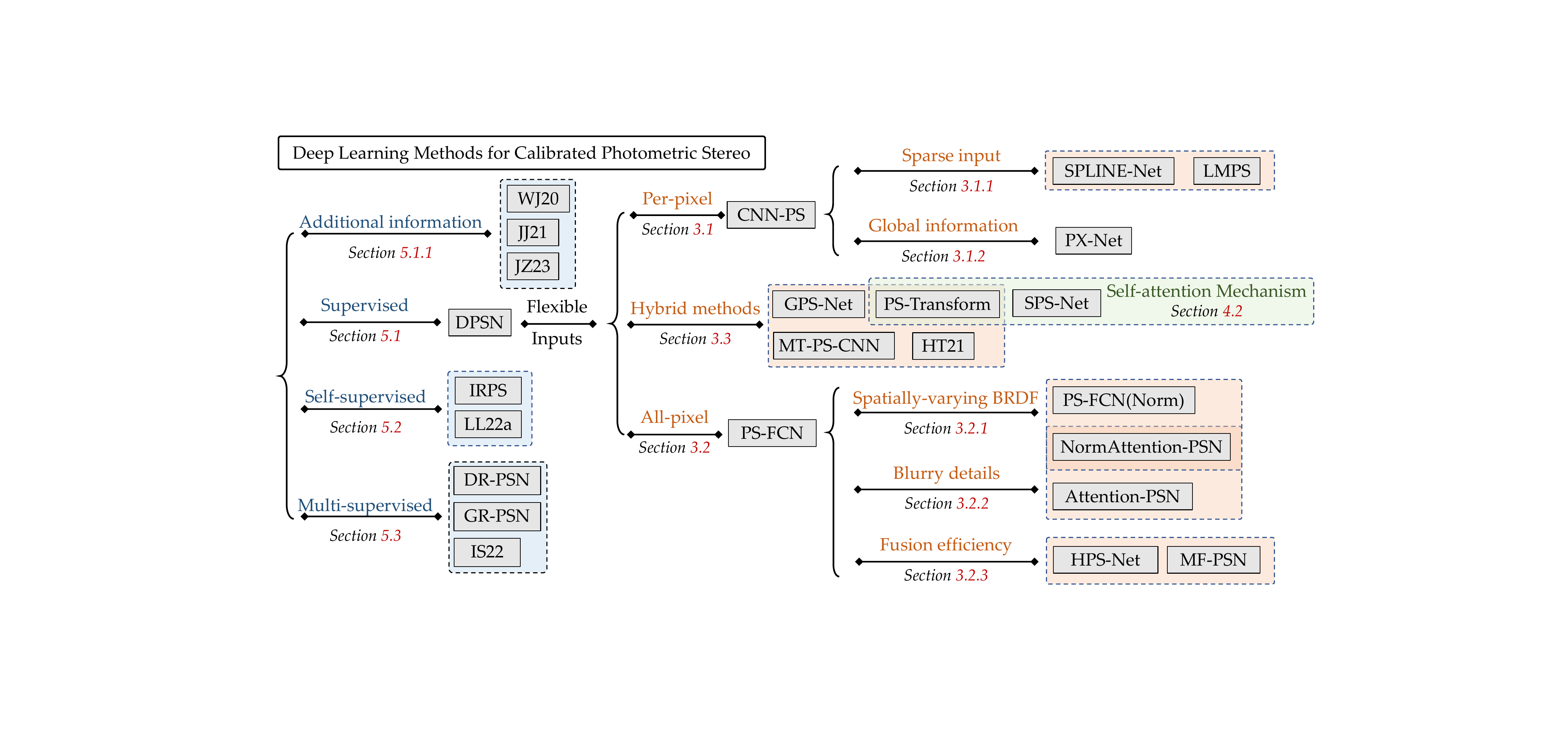}
  \end{center}
  \vspace*{-3mm}  \caption{Overview of the main deep learning methods for calibrated photometric stereo.}
  \label{overview}
    \vspace*{-3mm} 
\end{figure*}

However, real-world objects barely have the property of Lambertian reflectance. The non-Lambertian property of surfaces (as shown in the orange box in Fig. \ref{figshowps}) affects the proportional relationship of Eq. \ref{proportion}. Mathematically, we express the non-Lambertian property via the bidirectional reflectance distribution function (BRDF), depending on the material of the object. According to the previous taxonomies \cite{shi2019benchmark,santo2022deep,herbort2011introduction}, plenty of work has addressed non-Lambertian photometric stereo by modeling BRDF \cite{ikehata2014photometric,higo2010consensus,shi2014bi}, rejecting outlier regions \cite{wu2010robust,ikehata2012robust,wu2009photometric}, or setting exemplars \cite{hui2016shape,hertzmann2005example}. Nevertheless, designing appropriate reflectance models using general parametric BRDFs for photometric stereo is challenging, since these non-learning models tend to be accurate only for specific materials and often involve unstable optimization processes. In this context, early nonparametric attempts based on shallow artificial neural networks were introduced to establish a mapping between complex reflectance observations and surface normal \cite{iwahori1993neural,cheng2006neural,elizondo2008surface}. However, these models were restricted to limited materials that lack practical applications or require pre-training with a reference object with the same material as the target.

In 2017, DPSN \cite{santo2017deep} first attempted to use modern deep neural network architecture in the context of photometric stereo. It established the learning-based photometric stereo framework that more flexible mapping from reflectance observations to surface normal, breaking through the per-material-per-train limitation in early methods. DPSN \cite{santo2017deep} showed superior performance on non-Lambertian surfaces compared with traditional hand-crafted models which explicitly estimate the BRDF parameters and decouple the surface normals. However, this method required a fixed number and order of illumination directions during training and testing, which limited its generalization. 

To enhance the generalization, various deep learning-based approaches have been introduced.  This paper specifically concentrates on deep learning-based calibrated photometric stereo methods utilizing orthographic cameras and directional light sources. We categorize and summarize these methods from different perspectives, including input processing, supervision, and network architecture, the overview framework is shown in Fig. \ref{overview}. 

In this paper, we first categorize these deep learning-based calibrated photometric stereo methods based on how they process the input images, as per-pixel methods (\emph{i.e.}, by the observation map operation \cite{ikehata2018cnn} to record the intensity of each pixel) or all-pixel methods (\emph{i.e.}, by using aggregation model \cite{chen2018ps} to fuse whole patches). Different from the recent summary \cite{zheng2020summary} that only five calibrated learning-based photometric stereo models were listed, we comprehensively summarize and discuss the pros and cons of the various methods and how they evolve within these two categories. Additionally, we introduce a new classification, known as hybrid methods, which leverage both pixel- and patch-wise characteristics to enhance performance.

Second, the complexity and the number of parameters in learning-based models have significantly increased. Many advanced modules were integrated into surface-normal recovery tasks, such as ResNet \cite{he2016deep}, DenseNet \cite{huang2017densely}, HR-Net \cite{sun2019deep}, Transformer \cite{ashish2017attention}, \emph{etc}. Similarly, the selection of synthetic training data sets for photometric stereo became more diverse, \emph{i.e.}, rendering with MERL BRDF \cite{matusik2003data} or Disney’s principled BSDF \cite{mcauley2012practical}. In this paper, we also conduct a comprehensive summary and discussion of the network architectures and data sets utilized in previous deep learning-based photometric stereo methods.

In addition, we analyze the literature from the perspective of supervision, \emph{i.e.}, how 
the methods optimize the network (Section \ref{supvsec}). Most deep photometric stereo networks are trained with paired photometric stereo images (input) and surface normals (ground truths), \emph{i.e.}, supervised learning. Whether a photometric stereo network can be optimized in a self-supervised way? Whether additional information can be added to simplify the learning of surface-normal recovery? On these sides, this paper reviews recent attempts to expand and break through the supervised frameworks \cite{taniai2018neural,wang2020non,ju2023grpsn,ikehata2022does} and gives suggestions for future developments.

Based on the classifications and summaries provided above, we then evaluate more than 30 deep learning models for photometric stereo on the widely used benchmark \cite{shi2019benchmark} in dense input condition (Table \ref{benchcompare}) and sparse input condition (Table \ref{benchsparse}), respectively. We found that compared with traditional non-learning methods, deep learning-based photometric stereo models are superior in estimating surface normals. Finally, we point out the future trends in the field of photometric stereo. Our aim with this survey is to help researchers understand the state-of-the-art methods and position themselves to develop in this growing field, as well as highlight opportunities in future research. The project of this survey can be found in \url{https://github.com/Kelvin-Ju/Survey-DLCPS}.

\section{Problem Formulation}

Consider a pixel on a non-Lambertian surface with the normal $\boldsymbol{n}$ illuminated by a directional incident light $\boldsymbol{l}$. When a linear-response camera photographs this surface in the view direction $\boldsymbol{v}$, the pixel-measured intensity $m$ in image $\boldsymbol{M}$ can be approximated as follows:
\begin{equation}\label{imaging} 
m = \rho\left(\boldsymbol{n},\boldsymbol{l},\boldsymbol{v}\right)  \cdot \max \left\{\boldsymbol{n}^{\top}\boldsymbol{l}, 0\right\} + \epsilon, 
\end{equation}
where $\rho$ represents the BRDF, and $\max \left\{\boldsymbol{n}^{\top}\boldsymbol{l}, 0\right\}$ denotes the attached shadows, and $\epsilon$ represents global illumination effects (\emph{e.g.}, cast shadows and inter-reflections) and noise. Traditional photometric stereo methods computed the surface normals of general objects by solving the imaging model Eq. \ref{imaging} inversely, using more than three input images, but unknown BRDFs make the model difficult to fit (as shown in Fig. \ref{figshowps}). Similarly, deep learning-based calibrated photometric stereo methods aimed to learn a neural network model $f$ from $n$ different observations, as follows:
\begin{equation}\label{dlcps} 
f: \mathrm{Agg}(\boldsymbol{M^i}, \boldsymbol{l^i}) \to \boldsymbol{N}, i \in \{1, 2, \cdots, n\},
\end{equation}
where $f$ is the optimized deep neural network by the training data sets. Usually, the aggregation models (Agg) are determined by how they process the input images, such as observation maps, max-pooling models, or hybrid methods, which will be discussed in Section \ref{inputclass}. Most of the existing PS methods, \emph{i.e.}, calibrated photometric,  relied on having prior knowledge of the light directions and intensities for each image, while uncalibrated photometric stereo can estimate surface normals without lighting information. Note that the model $f$ becomes $f: \mathrm{Agg}(\boldsymbol{M^i}) \to \boldsymbol{N}, i \in \{1, 2, \cdots, n\}$ when addressing uncalibrated photometric stereo. Although uncalibrated photometric stereo has the advantage of not requiring pre-calibration of lighting conditions, it does face additional challenges because it needs to disentangle the lighting information from shading cues, making it a more complex problem to solve. In this paper, we mainly focus on the deep learning-based calibrated photometric stereo since it provides more universal frameworks and feature extraction models that can be extended to uncalibrated and other photometric stereo tasks. A brief discussion of the uncalibrated condition can also be found in Section \ref{uncalibrated} for a more comprehensive overview.

In the following subsections, we will discuss these deep learning-based calibrated photometric stereo methods from different perspectives.

\section{Categorization by Input Processing}
\label{inputclass}
The first deep learning method, DPSN \cite{santo2017deep}, made the order of illuminations and the number of input images unchanged, by a seven-layer fully-connected network. Therefore, the following methods focused on handling any number of input images with arbitrary light directions. In fact, this problem is equivalent to how to fuse a varying number of features in the networks. It is known that convolutional neural networks (CNNs) cannot handle a varying number of inputs during training and testing. Therefore, two approaches have been proposed in photometric stereo, \emph{i.e.}, to process the input images pixel-wise or patch-wise. Following the concept proposed in \cite{zheng2020summary}, we also call the pixel-wise and patch-wise processing methods as per-pixel methods (Section \ref{per}) and all-pixel methods (Section \ref{all}), respectively. We provide an in-depth summary of the development of these two approaches, in reference to the drawbacks of the initial methods (\emph{i.e.}, the observation map from CNN-PS \cite{ikehata2018cnn} and the max-pooling from PS-FCN \cite{chen2018ps}). In addition, we propose a new class, for hybrid methods (Section \ref{perall}), which fuse pixel- and patch-wise characteristics. As tabulated in Table \ref{algorithms}, we also summarize the algorithms and formulas of representative methods for each direction in Fig. \ref{overview}.

\subsection{Per-pixel methods}
\label{per}
\begin{figure}[t]
 \begin{center}
  \includegraphics[width=0.45\textwidth]{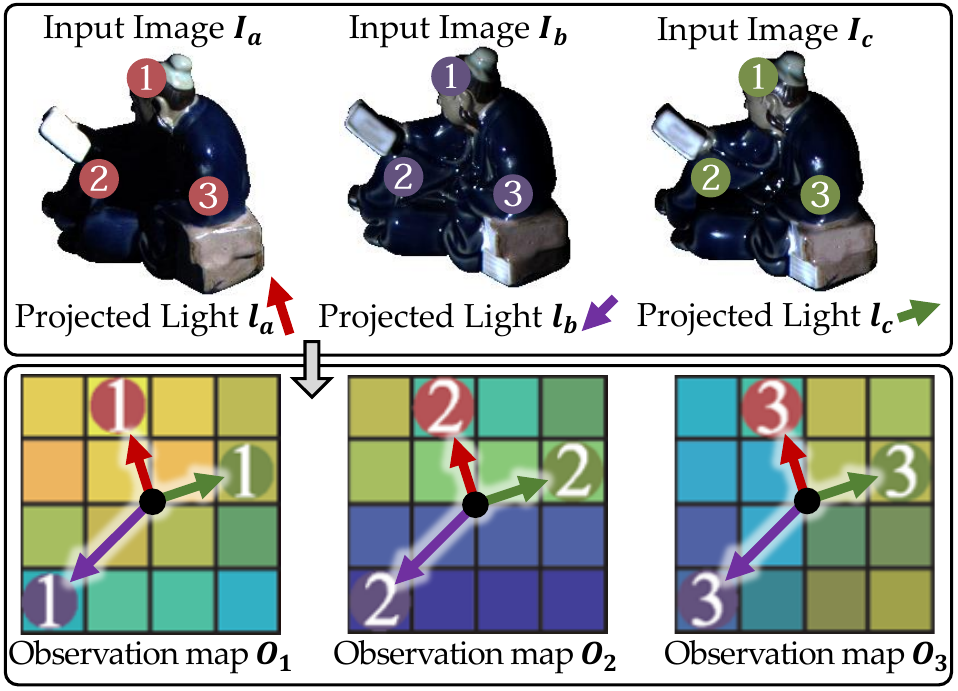}
  \vspace{-3mm} 
  \end{center}
  \caption{The illustration of the observation maps \cite{ikehata2018cnn}. Here, a, b, and c represent the number of input images (lights), while 1, 2, and 3 denote the index of pixel positions. }
  \vspace{-3mm} 
  \label{figobsmap}
\end{figure}

The per-pixel strategy was first implemented using the observation map in CNN-PS \cite{ikehata2018cnn}. The observation map essentially fused all observations pixel-by-pixel, capturing the inter-image intensity variations for each pixel. Observation maps were also widely used in recent near-field photometric stereo \cite{logothetis2023cnn} and multiview photometric stereo \cite{kaya2022uncertainty} tasks. Fig. \ref{figobsmap} illustrates the fusion rule, which is based on both the pixel intensity and the orthogonal projected light direction. Specifically, observation maps \cite{ikehata2018cnn} are determined by projecting light directions from a 3D space (hemisphere) onto a fixed-size observation map plane (along the axis-$z$ direction).  Each observation map can represent the feature at a single-pixel position. The observation map proves to be effective in photometric stereo for three reasons. First, its size is independent of the number of input images. Second, the values are independent of the order of the input images. Third, the information on the light directions and intensities is embedded in the observation map \cite{ikehata2018cnn}. Recently, Ikehata \cite{ikehata2022does} further took advantage of the physical interpretability of the observation map, making the observation map parse the physical intrinsic attributes to form a self-supervised inverse rendering pipeline.

\subsubsection{Problem of sparse input}
\label{sparseobs}
However, the observation map in the initial method \cite{ikehata2018cnn} also encounters some limitations. First, light directions are represented by unstructured vectors, while observation maps are grid data as images. When projecting a light vector onto a 2D coordinate system, the projected direction can not exactly correspond to the grid observation map. To improve the accuracy of projected light directions, the size of the observation map has to be large enough to approximately represent the unstructured projected vectors. Unfortunately, the number of input images (light directions) is sparse compared to the size of the observation map, which creates difficulties in extracting features. In fact, the sparse observation map affects network performance. The accuracy of CNN-PS drops significantly when inputting a small number of images (sparse condition), compared with the all-pixel methods.

\begin{figure}[!t]
 \begin{center}
  \includegraphics[width=0.36\textwidth]{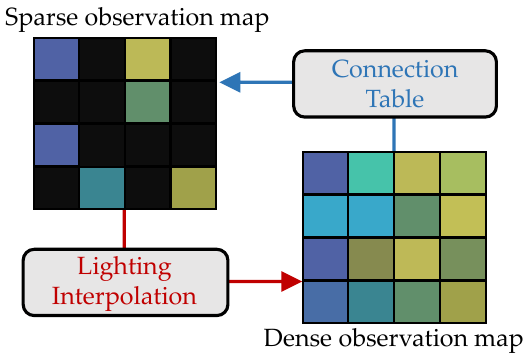}
  \end{center}
  \vspace{-3mm}
  \caption{ Per-pixel methods for sparse input images. SPLINE-Net \cite{zheng2019spline} uses the lighting interpolation network to generate dense observation maps, while LMPS \cite{li2019learning} applies the connection table used to select the most relevant illuminant directions in the sparse observation maps. }
  \label{figsparse}
  \vspace{-3mm} 
\end{figure}

In this regard, some works were proposed to solve the sparse input images problem, such as SPLINE-Net \cite{zheng2019spline} and LMPS \cite{li2019learning}. These two methods adopted opposite strategies to solve this problem. SPLINE-Net \cite{zheng2019spline} proposed a lighting interpolation network to generate dense lighting observation maps when the input was sparse (as shown in the red arrow of Fig. \ref{figsparse}). To optimize the lighting interpolation network and normal estimation network, SPLINE-Net further utilized a symmetric loss and an asymmetric loss to consider general BRDF properties explicitly and outlier rejections, respectively. On the other hand, LMPS \cite{li2019learning} reduced the demands on the number of images by only learning the critical illumination conditions. The method employed a connection table to select those illumination directions that were the most relevant to the surface normal prediction process (as shown in the blue arrow of Fig. \ref{figsparse}). Furthermore, a more thorough method \cite{yao2020gps} was to replace the structured observation map with an unstructured graph network, which will be introduced in Section \ref{perall}.

\subsubsection{Problem of global information}
\label{global}

On the other hand, since original per-pixel methods \cite{ikehata2018cnn} operate in isolation, which means the estimated normal vector of a surface pixel relies solely on the features extracted from that pixel itself, without leveraging information from adjacent pixels. As a result, it may lose the local context information of neighboring pixels when computing the feature map.

When the input observations exhibit deviations in photometric cues, per-pixel methods might exhibit reduced robustness compared to all-pixel methods, which consider all pixels in the input patch. For example, as mentioned in \cite{ikehata2018cnn}, where the first 20 images of the ``Bear'' object in the DiLiGenT benchmark data set \cite{shi2019benchmark} were less accurate: the intensity values around the bear's stomach region were lower than the adjacent regions, even though they should be higher due to specularities. When all 96 images of "Bear" were fed into the per-pixel method CNN-PS \cite{ikehata2018cnn}, the mean angular error increased dramatically, from 4.20 (with the first 20 images discarded) to 8.30, an increase of 97.62\%. In contrast, all-pixel methods demonstrated better robustness, \emph{e.g.}, with PS-FCN \cite{chen2018ps}, the error increased from 5.02 to 7.55, by only 50.40\%. This experiment illustrates the robustness of adjacent pixels.

To solve this limitation, some recent works incorporated global information into observation map-based per-pixel methods, which led to superior performance, such as PX-Net \cite{logothetis2021px}. PX-Net proposed an observation map-based method that considers global illumination effects, such as self-reflections, surface discontinuity, and ambient light, which enabled global information to be embedded in the per-pixel generation process. Additionally, PX-Net performed well in handling sparse conditions, in contrast to the original observation map-based method \cite{ikehata2018cnn}. Other methods, such as HT21 \cite{honzatko2021leveraging} and GPS-Net \cite{yao2020gps}, learned global information (intra-image features) by combining the per-pixel and all-pixel strategies. We will discuss these methods in Section \ref{perall}.

\subsection{All-pixel methods}
\label{all}

In contrast to per-pixel methods, which analyze each pixel individually in observations, all-pixel methods keep all the pixels together. All-pixel methods have the advantage of exploring intra-image intensity variations across an entire input image. The original all-pixel method was introduced in PS-FCN \cite{chen2018ps} through the use of a max-pooling layer, which operated in the channel dimension and fused features from an arbitrary number of inputs. At each position in the fused feature, the value was determined as the maximum among all the input features at that position. Consequently, this method allowed a convolutional network to work with features from any number of inputs. The max-pooling layer was inspired by aggregating multi-image information in other computer vision tasks \cite{wiles2017silnet,hartmann2017learned}. Compared to variable input methods like RNN \cite{graves2013speech}, the adopted max-pooling operation was order-agnostic, meaning it was not sensitive to the order in which the input images were provided. This attribute made it particularly suitable for photometric stereo. The all-pixel max-pooling operation offers several advantages. Firstly, it can handle an arbitrary number of input images without being affected by their order. Secondly, the use of whole image features includes valuable local context information, which enhances surface normals estimation. Thirdly, the patch-based input accelerates the training process compared to per-pixel methods. Lastly, all-pixel methods handle the input images and lighting directions (as extra information) separately, making them capable of predicting photometric stereo under unknown illuminations (uncalibrated photometric stereo).

\begin{figure}[t]
 \begin{center}
  \includegraphics[width=0.48\textwidth]{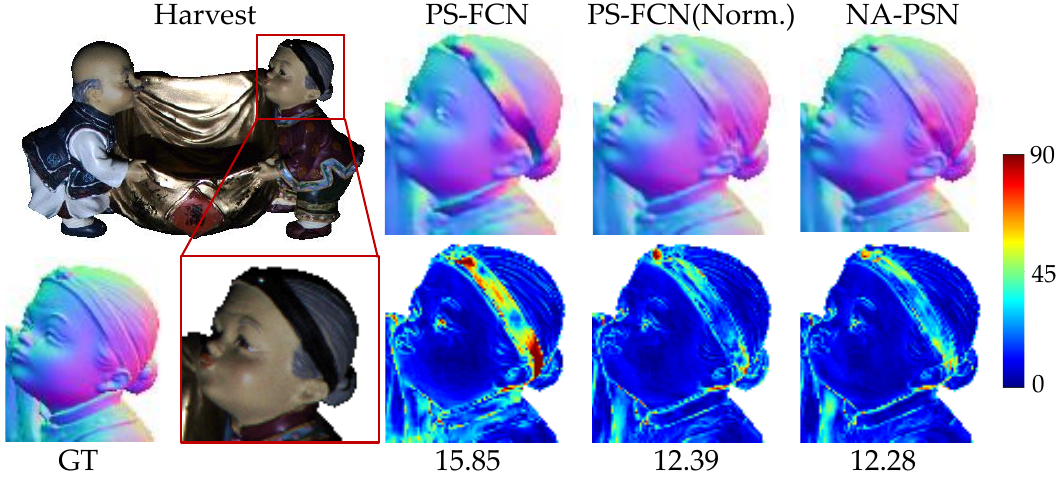}
  \end{center}
    \vspace{-4mm} 
  \caption{Examples of the predictions and error maps on spatially varying BRDF, from the object ``Harvest'' in the DiLiGenT data set \cite{shi2019benchmark}. NA-PSN is short for NormAttention-PSN. The numbers reveal the mean angular error in degrees. }
  \label{figsvbrdf}
    \vspace{-3mm} 
\end{figure}

\subsubsection{Problem of spatially varying BRDF }
\label{svall}

However, the original all-pixel method PS-FCN \cite{chen2018ps} had some limitations. To begin with, PS-FCN cannot handle surfaces with spatially varying materials. Since all-pixel methods leverage convolutional networks to process input in a patch-based manner, they may have difficulties in dealing with steep color changes caused by surfaces with spatially varying materials. It can be seen as the negative effect of considering observations in the neighborhood when computing the feature maps. As shown in Fig. \ref{figsvbrdf},  the head and collar region is with spatially varying BRDF. The original per-pixel method PS-FCN \cite{chen2018ps} was less effective in handling regions with spatially varying BRDF, where the color change of the beard influenced the surface normal map.
While improved methods, such as PS-FCN (Norm.) \cite{chen2022deep} and NormAttention-PSN \cite{ju2022Normattention}, showed significantly enhanced reconstruction results. This problem may be rooted in two key factors. Firstly, the feature extraction network encounters difficulty in decoupling the changes between the photometric shading cues and BRDFs. In other words, the feature extraction network may struggle to differentiate between changes in pixel values due to variations in surface structures and those resulting from different material properties. Secondly, the per-pixel methods inherently incorporate local context information, where each estimated surface normal vector depends on neighboring pixels when computing feature maps. Consequently, surface normal estimations can be influenced by spatially varying BRDF.

\begin{figure}[t]
 \begin{center}
  \includegraphics[width=0.44\textwidth]{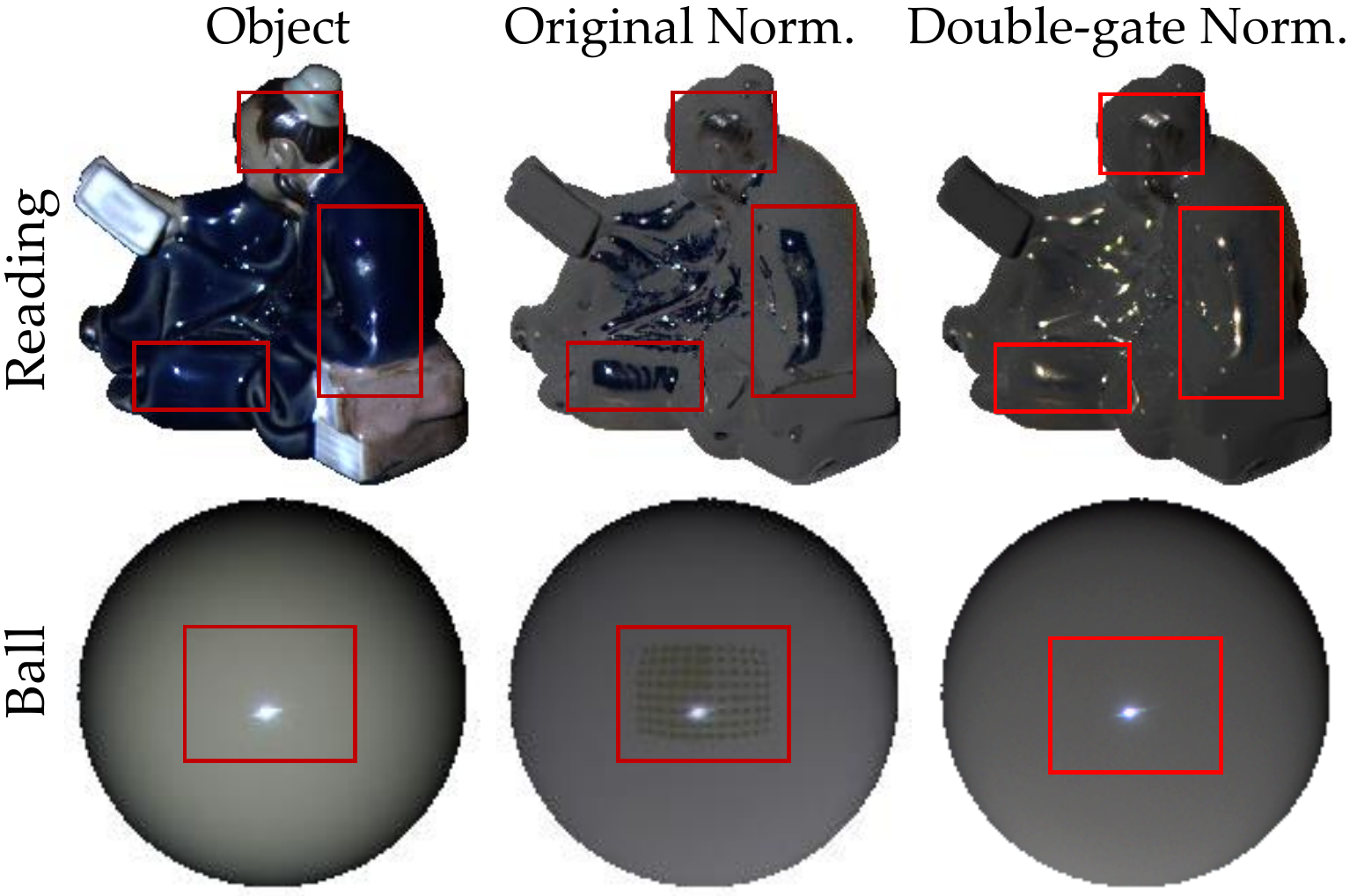}
  \end{center}
   \vspace{-3mm} 
  \caption{Comparison of the original normalization method \cite{chen2022deep} and the double-gate normalization method \cite{ju2022Normattention}, with the input object ``Reading'' and ``Ball''. The red boxes represent the regions exhibiting specular highlights. }
      \vspace{-2mm} 
  \label{fignormalization}
\end{figure}

To solve this limitation, Chen \emph{et al.} further proposed PS-FCN (Norm.) \cite{chen2022deep}. Rather than creating a large-scale training set with spatially varying materials, an observation normalization method, which concatenated all the observations and normalizes them, was introduced, as follows: 
\begin{equation}
\label{normalizationeq}
m'_i = \frac{m_{i}}{\sqrt{m_{1}^{2}+\cdots+m_{n}^{2}}} , \ i  \in \{1, 2,\cdots, n\},
\end{equation}
where $m_i$ and $m'_i $ represent the original and normalized pixel intensities in the $n$ images. Under the assumption of Lambertian reflectance, the effect of albedo can be removed. However, PS-FCN (Norm.) \cite{chen2022deep} cannot perfectly handle the condition of non-Lambertian surfaces. In regions with specular highlights, the denominator of Eq. \ref{normalizationeq} becomes larger, leading to the suppression of observations after normalization \cite{chen2022deep}. As shown in the red boxes in Fig. \ref{fignormalization},
the original normalization method \cite{chen2022deep} excessively suppressed the highlighted regions, whereas the double-gate normalization method \cite{ju2022Normattention} provided more reasonable shading cues in these regions. Although max-pooling can naturally ignore non-activated features, the suppressed observations are not equal to the suppressed features, \emph{i.e.}, the changing appearance of an observation may cause larger feature values. Therefore, Ju \emph{et al.} \cite{ju2022Normattention} proposed a double-gate observation normalization to better handle the
non-Lambertian surfaces with spatially varying materials. In the method, two gates were set at the lowest 10\% ($P_{10}$) and the highest 10\% ($P_{90}$) grayscale values of all pixels and put them on the denominator of Eq. \ref{normalizationeq}, as follows:
\begin{equation}
\label{nor3}
m'_i = \frac{m_{i}}{\sqrt{\sum_{k}m_k^2}}, \ k  \in \mathcal{S},
\end{equation}
where the set $\mathcal{S}$ is controlled by the two gates, such that
$m_i \in \mathcal{S}$ if $Gate(P_{10}) < m_i < Gate(P_{90})$, for $i = 1, 2, \cdots, n$. It can be seen that the non-Lambertian effects are removed in the red boxes in Fig. \ref{fignormalization}. However, this method has to concatenate with the original images, since discarding some grayscale values in the denominator can be viewed as a nonlinear process, which may affect the shading cues for photometric stereo \cite{ju2022Normattention}.

\subsubsection{Problem of blurry details}
\label{blurryall}

The second limitation of original all-pixel methods is that they may cause blurred reconstructions in complex-structured regions. We believe that the reasons mainly lie in three. (1) The convolutional models process patch-based input, which means that all normal points will mutually affect each other and cause blurring, especially in high-frequency areas. (2) The widely used Euclidean-based loss functions can hardly constrain the high-frequency (\emph{i.e.}, complex-structured) representations, because of the “regression-to-the-mean” problem \cite{isola2017image}, which results in blurry and over-smoothed images. (3) Previous network architectures pass the input through high-low-high resolutions, \emph{i.e.}, through an encoder-decoder architecture, which leads to the loss of prediction details and causes blurring.

\begin{figure}[!t]
 \begin{center}
  \includegraphics[width=0.42\textwidth]{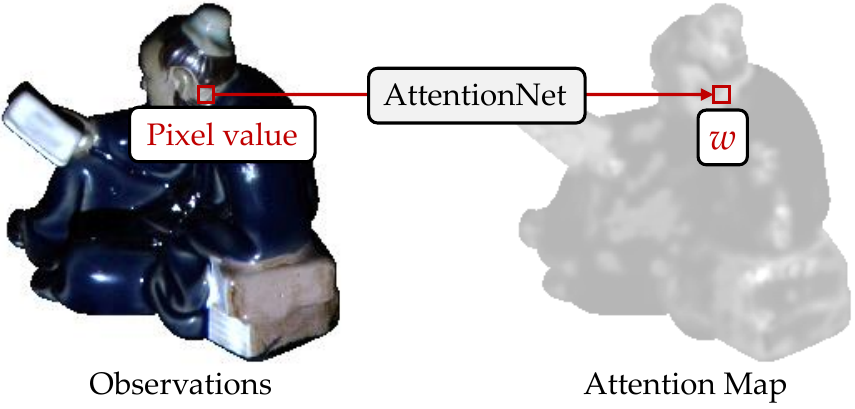}
  \end{center}
      \vspace{-3mm} 
  \caption{An example of an attention map from Attention-PSN \cite{ju2020pay}. $w$ is the weight of the pixel-wise attention-weighted loss (Eq. \ref{adaptiveloss}). }
  \label{figattentionpsn}
      \vspace{-3mm} 
\end{figure} 

In this regard, two different strategies were proposed to deal with the problem of blurred reconstruction in all-pixel methods. The first approach was to employ adaptive loss for different kinds of surfaces. Attention-PSN \cite{ju2020pay} was the first to propose an attention-weighted loss to produce detailed reconstructions, as follows:
\begin{equation}
\label{adaptiveloss}
\mathcal{L} = w\mathcal{L}_{\mathrm{gradient}} + (1- w) \mathcal{L}_{\mathrm{normal}}, 
\end{equation}
which learned a higher weight $(w)$ for the detail-preserving gradient loss $\mathcal{L}_{\mathrm{gradient}}$ and a lower weight $(1 - w)$ for the cosine loss $\mathcal{L}_{\mathrm{normal}}$ for high-frequency regions. As shown in Fig. \ref{figattentionpsn}, Attention-PSN \cite{ju2020pay} learned an attention map from input images, whose pixel values became the weights of the attention-weighted loss. However, the surface materials of an object may change rapidly in a flat or smooth region, which affects the gradient loss with a large weight in the region and dilutes the penalty on surface normals. Therefore, Ju \emph{et al.} further employed the above double-gate observation normalization to eliminate the influence of spatially varying surface materials, namely NormAttention-PSN \cite{ju2022Normattention}. 

On the other hand, the second approach was to preserve the high-resolution features via novel network architectures. CHR-PSN \cite{ju2022learning} proposed a parallel network structure for maintaining both deep features and high-resolution details of surface normals, inspired by the High-resolution Net (HR-Net) \cite{sun2019deep} for human pose estimation. Full-resolution features can always be preserved in the network, avoiding features passing layers from high to low resolution and blurring.

\subsubsection{Problem of fusion efficiency}
\label{max-poolingall}

The third is that the fusion mechanism of all-pixel methods, \emph{i.e.}, max-pooling, discards a large number of features from the input, reducing the utilization of information and affecting the estimation accuracy. Therefore, how to retain more features with key information is essential. Some methods \cite{yao2020gps,song2020photometric} fused max-pooling and average pooling via a concatenation operation. However, the improvement of adding average-pooling is limited because averaging features may smooth out saliency and dilute valuable features. Different from adding averaging information, Manifold-PSN \cite{ju2020learning} introduced nonlinear dimensionality reduction \cite{tenenbaum2000global} to convert features from high-dimensional feature spaces to low-dimensional manifolds. However, the manifold method truncated the backpropagation of the network. Therefore, the authors had to use the max-pooling layer to pre-train the extractor of the network, which was cumbersome and inefficient. 

On the other hand, some methods employed novel models to enhance feature fusion in their structures. MF-PSN \cite{liu2022deep} introduced a multi-feature fusion network, utilizing max-pooling operations at different feature levels in both shallow and deep layers to capture richer information. Besides, CHR-PSN \cite{ju2022learning}, SR-PSN \cite{ju2023estimating}, and MS-PS \cite{hardy2023multi} extended max-pooling at various scales with different receptive fields, rather than the depth. Furthermore, HPS-Net \cite{ju2023efficient} introduced a bilateral extraction module that generated positive and negative information before aggregation to better preserve useful data. Despite these advancements in feature fusion, none of these methods fully address the essential challenge of information loss, \emph{i.e.}, the max-pooling layer only extracts the maximum value, ignoring the rest.

Recently, the Transformer architecture \cite{ashish2017attention} has also been used to fuse and communicate features from different input images. PS-Transformer\cite{ikehata2021ps} first used a multi-head attention pooling \cite{lee2019set} to fuse an arbitrary number of input features. In this way, the number
of elements in a set was shrunk from an arbitrary dimension to one, by giving a learnable query $Q$ rather than only retaining the maximum value. Multi-head attention pooling \cite{lee2019set} can be seen as a global fusion method that considers all feature distributions, instead of only retaining the maximum value.

\subsubsection{Uncalibrated condition}
\label{uncalibrated}

Most of the existing methods, \emph{i.e.}, calibrated photometric stereo, require knowledge of the light direction and intensity for each image. However, calibrating the light involves complex operations and relies on specialized instruments, which may make it impractical for real-world applications. In contrast, uncalibrated photometric stereo can estimate surface normals without requiring lighting information. However, it encounters more challenges, such as the Generalized Bas-Relif (GBR) ambiguity \cite{belhumeur1999bas} and general non-Lambertian surface reflectance.

As discussed in Section \ref{per}, per-pixel methods rely on the projected light direction from 3D space onto the 2D observation map, where light directions are essential. Conversely, all-pixel methods handle input images and light directions separately. Therefore, the all-pixel strategy is first naturally applied in uncalibrated conditions. The per-pixel method PS-FCN \cite{chen2018ps} first addressed the uncalibrated problem by directly learning the mapping from input images to surface normals without concatenating light directions, denoted as UPS-FCN. However, the performance of UPS-FCN is far from satisfactory due to the complex interplay among shading cues, which include unknown lighting directions, surface normals, and reflectance properties. To address the uncalibrated case more effectively, most deep learning-based uncalibrated photometric stereo methods adopted a two-stage strategy. This involves first estimating the light directions and then estimating surface normals using both the estimated light information and input images, based on all-pixel networks \cite{chen2019self,chen2020learned,sarno2022neural,yang2022reddle,tiwari2022lerps,li2022self}.

SDPS-Net \cite{chen2019self} first proposed the two-stage deep learning architecture to reduce the learning difficulty in uncalibrated photometric stereo. It began by estimating light directions and intensities via the light calibration network, then applied an all-pixel-based normal estimation network to obtain the surface normal map. UPS-GCNet \cite{chen2020learned} used object shape and shading information as guidance to improve lighting estimation. Similarly, ReDDLE-Net \cite{yang2022reddle} incorporated diffuse and specular cues to enhance light estimation. Sarno \emph{et al.} \cite{sarno2022neural} employed differentiable neural architecture search (NAS) to automatically discover the most efficient neural architecture for both light calibration and normal estimation networks. In addition to supervised methods, a few uncalibrated methods were implemented in self-supervised and multi-supervised ways. For example, Kaya \emph{et al.} \cite{kaya2021uncalibrated} used an uncalibrated neural inverse rendering approach to handle unknown lights, and Li \emph{et al.} \cite{li2022self} allowed re-rendered errors to be back-propagated to the light sources and refined them jointly with the normals. Yang \emph{et al.} \cite{yang2022s} utilized the neural reflectance field to realize the 3D reconstruction from uncalibrated photometric stereo images with the capability of recovering invisible parts. Tiwari \emph{et al.} \cite{tiwari2022lerps} jointly trained the network with image relighting and used multiple loss functions to optimize the network.

\subsection{Hybrid methods}
\label{perall}

\begin{figure}[t]
 \begin{center}
  \includegraphics[width=0.48\textwidth]{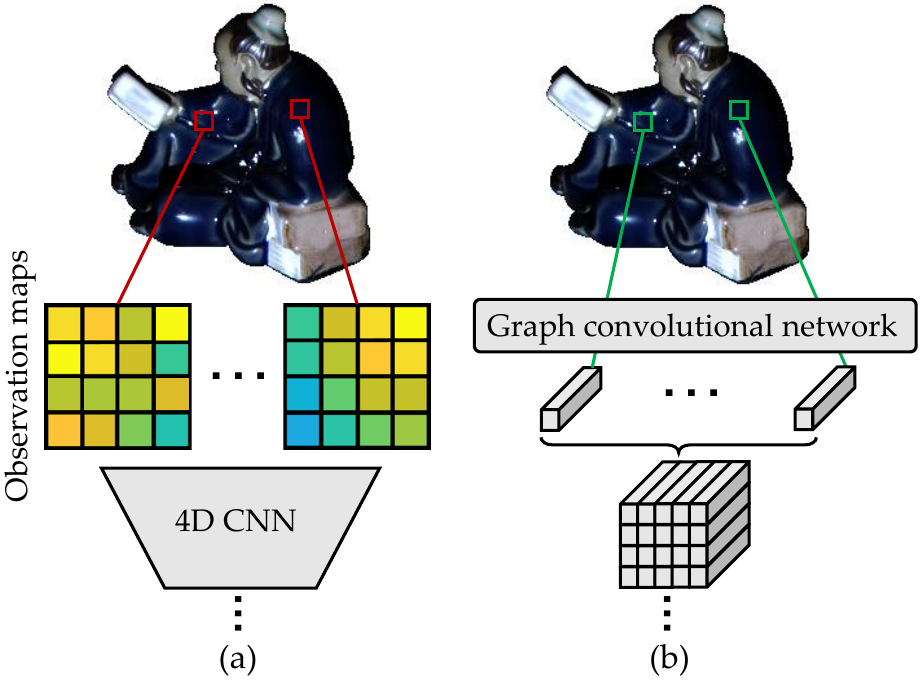}
      \vspace{-3mm} 
  \end{center}
  \caption{Hybrid methods for handling both per-pixel features and all-pixel features. (a) Schematic of the 4D CNN to capture the global effect of the observation maps used in HT21 \cite{honzatko2021leveraging}. (b) Schematic of the structure-aware Graph Convolution network to extract a fixed-size feature map, used in GPS-Net \cite{yao2020gps}.}
  \label{figmixed}
      \vspace{-2mm} 
\end{figure} 

As discussed above, both per-pixel and all-pixel methods come with their own sets of advantages and limitations. Per-pixel methods primarily focus on analyzing inter-image intensity variations at the pixel level. In contrast, all-pixel methods pay more attention to extracting features related to intra-image lighting variations. Hybrid approaches that combine these strategies may have the benefits of both per-pixel and all-pixel techniques.

In fact, the first mixed method can be found in learning-based multispectral photometric stereo\cite{ju2020dual}, which initially estimated a coarse surface normal map and subsequently refined it using a per-pixel approach, achieved through a fully connected network. Recently, MT-PS-CNN \cite{cao2022learning} proposed a two-stage photometric stereo model to construct inter-frame (per-pixel) and intra-frame (all-pixel) representations. Similarly, Yang \emph{et al.} \cite{Yang2023accurate} introduced a tandem manner for per-pixel and all-pixel feature extraction, namely PSMF-PSN. This network employed 3D convolutional layers to extract pixel-wise features. In addition, PS-Transformer \cite{ikehata2021ps} introduced a dual-branch feature extractor based on the self-attention mechanism \cite{ashish2017attention}, exploring both pixel- and image-wise features.  Honzatko \emph{et al.} \cite{honzatko2021leveraging} built upon the observation maps but incorporated spatial information using 2D and 4D separable convolutions to better capture global effects. Differently, GPS-Net \cite{yao2020gps} introduced a structure-aware graph convolutional network \cite{chang2018structure} to establish connections between an arbitrary number of observations per pixel, without relying on observation maps. Subsequently, convolutional layers were employed to extract spatial information. These hybrid methods may benefit from per-pixel and all-pixel approaches. As shown in Fig. \ref{figmixed}, we summarize the hybrid strategy of HT21 \cite{honzatko2021leveraging} and GPS-Net \cite{yao2020gps}.

However, all existing hybrid methods follow a sequential and independent approach to extracting per-pixel and all-pixel features.  Future research may focus on effective ways of combining these two feature types and consider the learning process as a holistic approach, rather than treating it as two separate stages.

\section{Network Architectures}
\label{archit}

With the development of deep learning techniques, deep learning-based photometric stereo networks have used many advanced modules. In this Section, we will review these modules and compare their advantages and drawbacks in the task of surface normal recovery.

\subsection{Convolutional networks}

\begin{figure*}
 \begin{center}
  \includegraphics[width=0.92\textwidth]{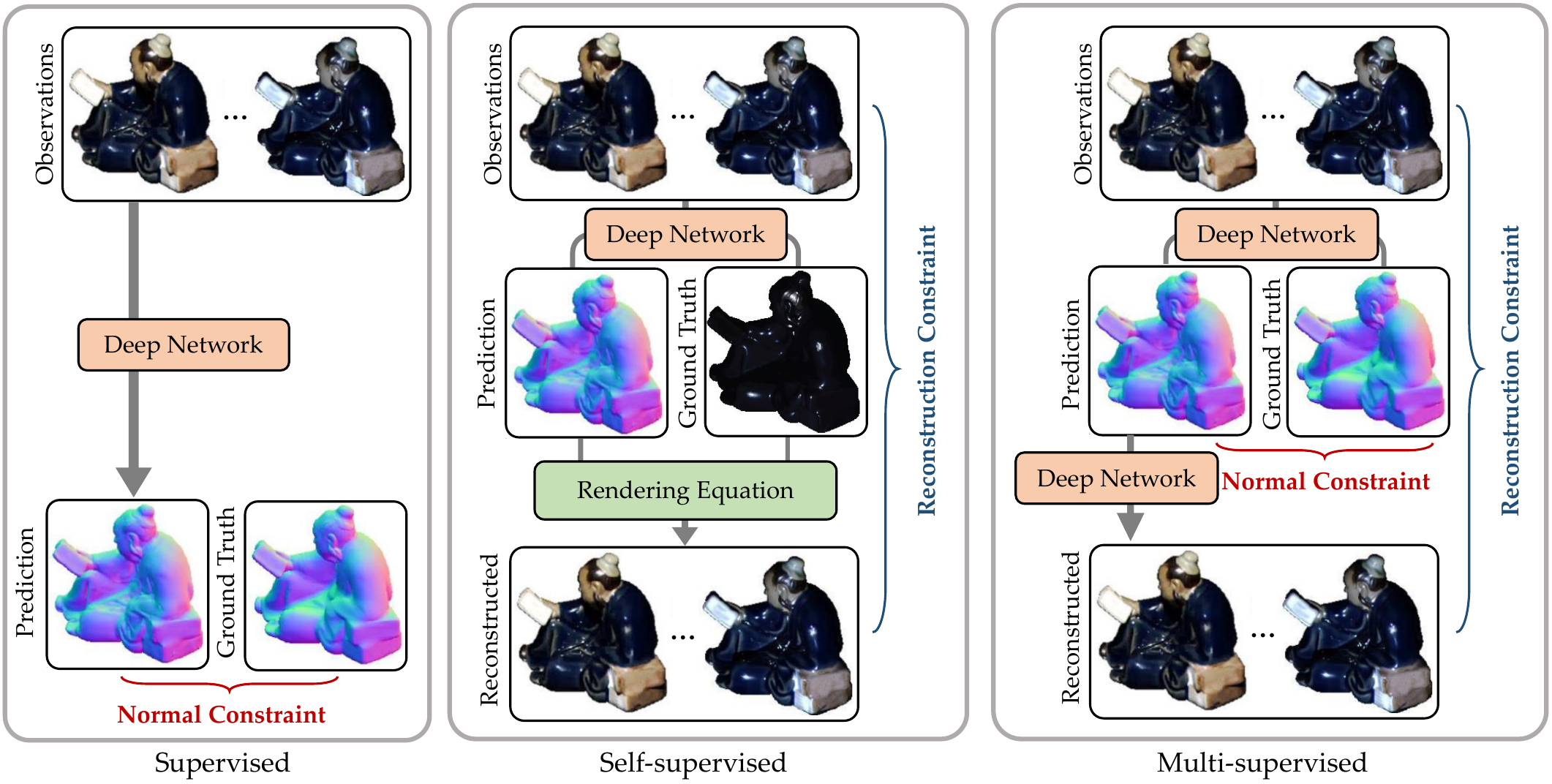}
  \end{center}
      \vspace{-3mm} 
  \caption{Comparison of the frameworks of supervised, self-supervised, and multi-supervised photometric stereo.}
  \label{figsupervision}
      \vspace{-4mm} 
\end{figure*}

In the beginning, DPSN \cite{santo2017deep,santo2022deep} utilized Multilayer Perceptron (\emph{i.e.}, fully connected layers) with dropout layers to map the surface normals from observations pixel by pixel. However, this architecture ignores adjacent information and cannot handle a flexible number of input images. Therefore, PS-FCN \cite{chen2018ps} and CNN-PS \cite{ikehata2018cnn} were proposed to handle an arbitrary number of input images by different strategies (max-pooling and observation map). PS-FCN \cite{chen2018ps} applied a fully convolutional plain network to learn surface normals, while CNN-PS \cite{ikehata2018cnn} used a variant of the DenseNet architecture \cite{huang2017densely} to estimate surface normals from an observation map. The DenseNet architecture \cite{huang2017densely} has been widely used in subsequent networks, such as LMPS \cite{li2019learning}, SPLINE-Net \cite{zheng2019spline}, MF-PSN \cite{liu2022deep}, and PX-Net \cite{logothetis2021px}, due to its excellent feature extraction capacity. Similarly, ResNet \cite{he2016deep} was also widely used in deep learning-based photometric stereo methods\cite{ju2020pay, ju2020learning,yao2020gps,ju2021incorporating}, which can effectively avoid gradient vanishing in deep networks. However, the above structures ignore keeping the high resolution of the features, \emph{i.e.}, passing the features sequentially from high-to-low resolution layers, and then increasing the resolution. This operation is suitable for a high-level task that needs semantic features. However, it may cause information loss and blurring for the per-pixel prediction photometric stereo task. Therefore, some works \cite{ju2022learning,ju2022Normattention} introduced a parallel multi-scale structure, inspired by the improvement of HR-Net \cite{sun2019deep} in the human pose estimation task. HR-Net \cite{sun2019deep} employed a parallel network structure to extract features at three scales, avoiding the feature map being changed from low resolution to high resolution, where the feature extraction process maintained both the deep features with high semantic and high-resolution features having details for surface-normal prediction. 

\subsection{Self-attention mechanism}
\label{selfatt}
Transformer with a self-attention module \cite{ashish2017attention} was first proposed in the field of natural language processing. It has also been widely used in many computer vision tasks, where self-attention was employed in the spatial dimensions to capture non-local feature dependencies. Recently, two works \cite{liu2020sps,ikehata2021ps} introduced the self-attention mechanism to aggregate features under different lights in the context of photometric stereo. SPS-Net \cite{liu2020sps} was the first to propose a self-attention photometric stereo network, which aggregated photometric information through a self-attention mechanism. Ikehata \emph{et al.} \cite{ikehata2021ps} then presented PS-Transformer, which uses the self-attention mechanism to capture complex interactions in sparse photometric stereo. PS-Transformer \cite{ikehata2021ps} designed a dual branch to explore pixel and image-wise features. Therefore, intra-image spatial features and inter-image photometric features are better extracted than with SPS-Net \cite{liu2020sps}. Recently, Ikehata introduced two photometric stereo methods: UniPS \cite{ikehata2022universal} and SDM-UniPS \cite{ikehata2023scalable}. These methods can handle natural lighting conditions by learning global lighting contexts from individual images by interacting with others, discarding assuming specific lighting models. In these approaches, the self-attention model \cite{ashish2017attention} served as the backbone to facilitate non-local interactions and as the aggregation method to fuse arbitrary features.

\subsubsection{Discussion}

The Transformer module showed significant performance improvements in other computer vision domains \cite{bertasius2021space,strudel2021segmenter}. Similarly, the photometric stereo task can leverage the self-attention module effectively. Theoretically, the surface normal of a point only depends on itself, rather than its relationship with distant points. However, due to the presence of shadows and inter-reflections, capturing long-range context becomes essential for accurate feature extraction. Therefore, Transformer-based photometric stereo models can benefit from both the non-local information acquired through the self-attention module and the embedded local context information obtained through traditional convolutional layers. Furthermore, the effectiveness of the Transformer \cite{ikehata2021ps,ikehata2022universal,ikehata2023scalable} can facilitate communication and aggregation of features flexibly, by using multi-head attention pooling \cite{lee2019set}.

However, Transformer-based photometric stereo methods face some limitations. The Transformer module has greater modeling flexibility and can focus on the information at any position. Consequently, in contrast to convolutional networks, it requires larger-scale training data sets. Furthermore, it is widely recognized that the Transformer model imposes substantial computational demands, especially when dealing with a large number of elements \cite{ikehata2021ps}. Hence, future research should explore the adaptation of Transformer-based photometric stereo methods to address dense problems more efficiently.

\section{Categorization by Supervision}
\label{supvsec}

As a mapping task, conventional learning-based photometric stereo methods optimized the network by minimizing the distance between predicted surface normals and ground-truth surface normals \cite{santo2017deep,ikehata2018cnn,chen2018ps}, supervised by pairs of photometric stereo images and their surface normals. However, learning-based 3D tasks face challenges due to the difficulties in acquiring and aligning a large number of ground truths. To solve this issue, some researchers have investigated 
self-supervised learning in photometric stereo \cite{taniai2018neural,kaya2021uncalibrated,li2022neural}. Besides, many works further improved the performance by introducing additional supervision \cite{ju2021recovering,ju2023grpsn,ikehata2022does} or additional information to simplify optimization \cite{wang2020non,ju2021incorporating,ju2023learning}. In Fig. \ref{figsupervision}, we summarize the differences among supervised, self-supervised, and multi-supervised photometric stereo networks. 

\subsection{Supervised photometric stereo methods}
\label{supervised}

Plenty of deep photometric stereo networks have been proposed with improved performance, compared to traditional handcrafted photometric stereo methods. These learning-based models show the potential ability of deep neural networks with supervised optimization, \emph{i.e}, a large amount of data with ground-truth surface normals during the training stage. Among these supervised models, some methods \cite{santo2017deep,santo2022deep,ikehata2018cnn,li2019learning} utilized the L2 loss (\emph{i.e.}, mean squared error loss), as follows:
\begin{equation}
\label{l2loss}
\mathcal{L} = \lVert  \boldsymbol{n_p} - \boldsymbol{\tilde{n}_p}  \rVert^2_2, 
\end{equation}
while more methods applied cosine similarity loss, as follows:
\begin{equation}
\label{cosloss}
\mathcal{L} = 1 - \boldsymbol{n_p}\odot\boldsymbol{\tilde{n}_p} , 
\end{equation}
where $\odot$ represents the dot-product operation. In this case, $\boldsymbol{n_p}\odot\boldsymbol{\tilde{n}_p}$ will be close to 1 when the predicted $\boldsymbol{\tilde{n}_p}$ is similar to the ground truth $\boldsymbol{n_p}$, and Eq. (\ref{cosloss}) will approach 0. Intuitively, the cosine similarity loss is more suitable for surface-normal estimation, because it directly measures the difference in orientation between two vectors. However, no evidence from previous work shows that L2 loss reduces the accuracy of estimated surface normals with the same network architecture and settings. 

\subsubsection{Additional information}
\label{additional}

Recently, some supervised photometric stereo networks improved performance with additional information to make optimization more efficient \cite{wang2020non,ju2021incorporating,ju2023learning}. The additional information can be regarded as prior knowledge used to simplify the optimization of deep networks through weight parametrization. In contrast to previous deep learning approaches that solely derived the normal space from the observed shading cues, these methods leveraged both additional information and observations to learn the surface normal. Consequently, these methods had the capacity to reduce the learning hypothesis space, leading to easier feature extraction, faster convergence, and improved learning accuracy.

Wang \emph{et al.} \cite{wang2020non} proposed a non-Lambertian photometric stereo network with the additional collocated light image. Their model leveraged the monotonicity of isotropic reflectance and the univariate property of the supplementary collocated light to facilitate the decoupling of the surface normal from the reflectance function, in conjunction with the input photometric stereo images. Ju \emph{et al.} \cite{ju2021incorporating} incorporated initial normal priors to enhance the accuracy of surface normal predictions for objects. This approach relied on prior surface normals based on Lambertian assumption \cite{Woodham1980Photometric} to reparameterize network weights, enabling the alignment of mappings in the same normal space and increasing the focus on the errors in the prior normal. Similarly, Ju \emph{et al.} \cite{ju2023learning} proposed an additional reflectance-guided photometric stereo network, which employed a dual-branch extractor to combine information from both prior reflectance and photometric stereo images. Furthermore, the inclusion of prior reflectance helped eliminate the impacts of surfaces with spatially varying reflectance for photometric stereo methods. These methods can enhance performance by incorporating additional information to streamline the optimization process.

In general, supervised photometric stereo methods can achieve superior performance, but these methods are limited due to the difficulties in acquiring accurate ground truth for the photorealistic training sets, and there is a gap between real photo images and synthetic images due to rendering techniques.

\subsection{Self-supervised photometric stereo methods}
\label{self-supervised}

As discussed above, deep learning techniques drastically advanced the photometric stereo task. Current existing deep learning methods usually solve the problem in a supervised training manner. These methods relied on a large amount of training data with ground truth. However, measuring the surface normals of real objects is very difficult and expensive, because it needs high-precision 3D scanners to reconstruct the ground-truth shape, and requires much manpower to align the viewpoints between surface normal maps and multiple images (pixel alignment). Until now, only three real-world scene data sets with ground truth have been proposed \cite{shi2019benchmark,ren2022diligent102,wang2023diligent}. However, these data sets only contained 10 to 100 object scenes and were far from being utilized for training a modern deep neural network. Synthetic training data is a possible way \cite{santo2017deep, chen2018ps,ikehata2018cnn}, but synthetic images should account for various realistic BRDF, object shapes, cast shadows, and inter-reflections, \emph{etc}. Existing BRDF databases \cite{matusik2003data,mcauley2012practical} and renderers still required efforts to generate photo-realistic synthetic images.

\begin{table*}[t]
\vspace*{-1mm}
\begin{center}
\caption{Representative calibrated deep learning photometric stereo algorithms and formulations for each taxonomy.} 
\vspace*{-4mm}
\label{algorithms}
\renewcommand\arraystretch{1.4}
\scalebox{0.82}{
\begin{tabular}{ | c | c| c | c |}

\hline
Taxonomy&Sec.& Method & Formulation \\
\hline
Fixed input & \ref{inputclass}&DPSN \cite{santo2017deep}& $\boldsymbol{n_p}=f (\tilde{m}^{i}_p)$, where $\tilde{m}^{i}_p$ means the order and the number of the inputs are fixed.   \\
\hline
Per-pixel input&\ref{per}&CNN-PS \cite{ikehata2018cnn} & $\boldsymbol{n_p}= f ( \mathrm{obs} (m^i_p, l^i ))$, please see $\mathrm{obs}$ in Section \ref{per}. \\
\hline
Sparse input&\ref{sparseobs}&LMPS \cite{li2019learning} & $\boldsymbol{n_p}= f (D(\mathrm{obs} (m^i_p, l^i )))$, where $D$ stands for the connection table. \\
\hline
Global information&\ref{global}&PX-Net \cite{logothetis2021px} & $\boldsymbol{n_p}= f (\mathrm{obs} (G(m^i_p), l^i ))$, where $G$ stands for the global illumination effects. \\
\hline
All-pixe input&\ref{all}&PS-FCN \cite{chen2018ps} & $\boldsymbol{N}=f_R( \mathrm{max} \{ f_E (\boldsymbol{M^i}, l^i)\})$, please see $\mathrm{max}$ in Section \ref{all}, $f_E$ and $f_R$ mean the Extractor and Regressor. \\
\hline
Spatially varying BRDFs&\ref{svall}&PS-FCN (Norm.) \cite{chen2022deep} &  $\boldsymbol{N}=f_R( \mathrm{max} \{ f_E ( \mathrm{norm} (\sum_{i}^{n}(\boldsymbol{M^i})), l^i)\})$, where $\mathrm{norm}$ stands for the Observation Normalization. \\
\hline
Blurry details&\ref{blurryall}&Attention-PSN \cite{ju2020pay} & $\boldsymbol{N}=f_R( \mathrm{max} \{ f_E (\boldsymbol{M^i}, l^i)\})$, with minimizing the adaptive loss $\mathcal{L} = \lambda \mathcal{L}_{\mathrm{gradient}} + (1- \lambda) \mathcal{L}_{\mathrm{normal}}$.\\
\hline
Fusion efficiency&\ref{max-poolingall}&MF-PSN \cite{liu2022deep} &$\boldsymbol{N}=f_R( \mathrm{max}_d \{ \mathrm{max}_s \{f_E (\boldsymbol{M^i}, l^i)\}, f_E (\boldsymbol{M^i}, l^i) \})$, where $s$ and $d$ mean shallow and deep, respectively.\\
\hline
Hybrid input&\ref{perall}&GPS-Net \cite{yao2020gps}& $\boldsymbol{N}=f(\cup_{p}^{H \times W} \mathrm{gcn} (m^i_p, l^i))$, where $\mathrm{gcn}$ stands for the Structure-aware Graph Convolution filters.\\
\hline
Self-attention&\ref{selfatt}&PS-Transformer\cite{ikehata2021ps} & $\boldsymbol{N}= f_R (f_{E}(m_p^i, l^i), f_{E} (\boldsymbol{M^i}))$, where $f_{E}$ and $f_{R}$ are layers with the Self-attention Mechanism.
\\
\hline
Additional information&\ref{additional}&WZ20 \cite{wang2020non} & $\boldsymbol{n_p}=  f_R( \mathrm{max} \{ f_E (m^i_p, l^i, m^0_p\})$, where $m^0_p$ stands for the collocated light observation. \\
\hline
Self-supervised&\ref{self-supervised}&IRPS \cite{taniai2018neural} & $\boldsymbol{N}= f_P(\boldsymbol{M^i})= f_I ^{-1} (\boldsymbol{M^i})$, with minimizing the self-supervised loss $ \mathcal{L} =f_I(f_P(\boldsymbol{M^i}),l^i) -\boldsymbol{M^i} $.\\

\hline
Multi-supervised&\ref{mutlisupervisions}&DR-PSN \cite{ju2021recovering} & $\boldsymbol{N}= f_P(\boldsymbol{M^i},l^i )= f_I ^{-1} (\boldsymbol{M^i})$, where $f_P$ and $f_I$ mean the Normal regression and Dual regression.
\\
\hline

\end{tabular}}
\end{center}
\vspace*{-4mm}
\end{table*}

To overcome the above shortcomings, some researchers introduced the self-supervised learning strategy, which only needs photometric stereo images, rather than pairs with ground truth surface normals \cite{taniai2018neural,kaya2021uncalibrated,li2022neural}. The pipeline of the self-supervised photometric stereo methods can be described as Eq. \ref{eqself}, as follows:
\begin{equation}
\label{eqself}
\boldsymbol{M}_{R}= \Psi (\Phi(\boldsymbol{M})),
\end{equation}
where $\Phi$ represents the model responsible for extracting the surface normal, while $\Psi$ denotes the method used to re-render the reconstructed images $\boldsymbol{M}_{R}$. In this case, the estimated surface normal $ \boldsymbol{\tilde{N}} = \Phi (\boldsymbol{M})$ is optimized by minimizing the reconstruction loss (\emph{e.g.}, L2 loss) between the input images $\boldsymbol{M}$ and the re-rendered images $\boldsymbol{M}_{R}$, without requiring the ground-truth surface normal of $\boldsymbol{M}$.

Taniai and Maehara \cite{taniai2018neural} first proposed a self-supervised convolutional network
that took the whole set of images as input, namely IRPS. The model directly generated surface normals by minimizing the reconstruction loss between re-rendered images obtained via the rendering equation and input images. Furthermore, IRPS \cite{taniai2018neural} avoided the unfixed number of input photometric images through its physics-based rendering approach. However, IRPS suffered from expensive computation \cite{zheng2020summary} and failed to model inter-reflections. Moreover, its loss function was not robust and susceptible to noise \cite{kaya2021uncalibrated} because the surface normals were initialized by using the Lambertian assumption \cite{Woodham1980Photometric}. Therefore, IRPS was further extended by Kaya \emph{et al.} \cite{kaya2021uncalibrated} to deal with inter-reflection by explicitly modeling the concave and convex parts of a complex surface. However, both \cite{taniai2018neural,kaya2021uncalibrated} implicitly
encoded the specular components as features of the network and
fail to consider shadows in the rendering process. To solve the limitations, Li \emph{et al.} \cite{li2022neural} proposed a coordinate-based deep network to parameterize the unknown surface normal and the unknown reflectance at every surface point. The method learned a series of neural specular basis functions to fit the observed specularities and explicitly parameterized shadowed regions by tracing the estimated depth map. However, the method may fail in the presence of strong inter-reflections. 

In summary, self-supervised photometric stereo models alleviate the demand for extensive 3D data sets. However, these self-supervised models face computational burdens due to their relatively large parameter sizes, which may restrict their applicability in industrial settings. We envision that future progress in self-supervised photometric stereo will involve the refinement of rendering equations for increased accuracy, the creation of more lightweight models, and the enhancement of the efficiency of reconstruction loss.

\subsection{Multi-supervised photometric stereo methods}
\label{mutlisupervisions}

Previous research \cite{wang2020non,ju2021incorporating} demonstrated improved performance by incorporating additional input information within supervised learning frameworks. Another approach to simplifying the learning process is to introduce more forms of supervision. In this paper, we refer to methods that utilize multiple forms of supervision as ``Multi-supervised'' \cite{ju2021recovering,ikehata2022does,ju2023grpsn}.

In this context, Ikehata \cite{ikehata2022does} proposed a network to deconstruct the observation map into physical interpretable components such as surface normal, surface roughness, and surface base color. These components were then integrated via the physical formation model \cite{burley2012physically}. Consequently, the training loss for optimization consisted of the normal reconstruction loss and the inverse rendering loss. On the other hand, Ju \emph{et al.} \cite{ju2021recovering} introduced a dual regression network for calibrated photometric stereo, known as DR-PSN. This network combined the surface-normal constraint with the constraint of the reconstructed re-lit image. Additionally, GR-PSN \cite{ju2023grpsn} utilized a parallel framework to simultaneously learn two arbitrary materials for an object and included an additional material transform loss. These methods employed an inverse subnetwork to re-render reconstructed images based on predicted surface normals. In contrast to previous inverse rendering methods \cite{taniai2018neural,kaya2021uncalibrated,li2022neural,ikehata2022does}, DR-PSN and GR-PSN used CNNs to render reconstructed images rather than following the rendering equation.

Finally, based on the taxonomy discussed in Sections \ref{inputclass}, \ref{archit}, and \ref{supvsec}, we formulate these representative deep learning-based calibrated photometric stereo methods in Table \ref{algorithms}. In these formulas, the predicted surface normals are represented by $\boldsymbol{N}$ from input photometric images $\boldsymbol{M^i}$ or $\boldsymbol{n_p}$ from pixel $m^i_p$, where $p$  stands for the index of spatial resolution $H \times W$, $i  \in \{1, 2,\cdots, n\}$ stands for the index of inputs. $f$ represents deep neural networks for learning surface normals.

\begin{table*}[!htp]
\begin{center}
\vspace*{-1mm}
\caption{Summary of data sets for deep learning photometric stereo.} 
\vspace*{-4mm}
\label{data settable}
\renewcommand\arraystretch{1.3}  
\scalebox{0.98}{
\begin{tabular}{ | c| c| c| c| c| c| }
\hline
Data set & BRDF & Ground Truth & Number of Sample & Trained methods \\
\hline
Blobby and Sculpture \cite{chen2018ps}& MERL \cite{matusik2003data}, homogeneous& \multirow{2}{*}{Synthetic normal} &85212&\makecell[c]{ \cite{chen2018ps,chen2022deep,li2019learning,ju2020learning,ju2020pay,cao2022learning,ju2021incorporating}\\ \cite{ju2021recovering ,ju2022learning,ju2022Normattention,yao2020gps,liu2022deep,liu2020sps,wang2020non}}\\
\cline{0-1} \cline{4-5}
CyclePS \cite{ikehata2018cnn}& Disney \cite{mcauley2012practical}, spatially varying && 45 (75 in \cite{ikehata2021ps})& \cite{ikehata2018cnn,ikehata2021ps,zheng2019spline,logothetis2021px,honzatko2021leveraging} \\
\hline
Gourd \& Apple \cite{alldrin2008photometric}&\multirow{4}{*}{Real object, spatially varying} & \multirow{2}{*}{Not provide} & 3 & \multirow{5}{*}{Test sets}\\
\cline{0-0}\cline{4-4}
Light Stage Data Gallery \cite{goldman2009shape} & &  & 9 & \\
\cline{0-0}\cline{3-4}
DiLiGenT \cite{shi2019benchmark} &  & \multirow{2}{*}{3D scanner} &10 &  \\
\cline{0-0}\cline{4-4}DiLiGenT-$\Pi$ \cite{wang2023diligent} & &&30&\\
\cline{0-3}
DiLiGenT$10^2$ \cite{ren2022diligent102} & Real object, homogeneous & CAD + CNC & 100 & \\

\hline
\end{tabular}}
\end{center}
\vspace*{-5mm}
\end{table*}

\section{Data Sets of Photometric Stereo}
\label{data setsec}

The training and testing of supervised photometric stereo networks require the ground truth normal maps of objects. However, obtaining ground-truth normal maps of real objects is a difficult and time-consuming task. Although many data sets have been established in other 3D reconstruction tasks \cite{jensen2014large,chang2015shapenet,aanaes2016large}, most of their objects were simple in reflectance and shape, and the number of different lighting conditions was small \cite{shi2019benchmark}. This section will review data sets for deep learning-based photometric stereo methods and summarize them in Table \ref{data settable}. It is worth noting that we mainly review these data sets for directional lighting photometric stereo methods. Those for multiview photometric stereo \cite{li2020multi}, near-field photometric stereo \cite{mecca2021luces}, and multispectral photometric stereo \cite{ju2018demultiplexing} are not discussed here.

\subsection{Training data sets}

Training a deep photometric stereo network needs to render plenty of materials, geometries, and illumination. Researchers have to establish synthetic training data sets by rendering 3D shapes with different reflectance. There are two mainstream data sets. 

\subsubsection{Blobby and Sculpture data set}

The Blobby and Sculpture data set was proposed in DPSN \cite{santo2017deep} and improved in PS-FCN \cite{chen2018ps}. PS-FCN applied 3D shape models from the Blobby shape data set \cite{johnson2011shape} and the Sculpture shape data set \cite{wiles2017silnet}, as well as the MERL BRDF data set \cite{matusik2003data} to provide surface reflectance. The Blobby shape data set contained ten objects with various shapes. The Sculpture shape data set \cite{wiles2017silnet} further provided more complex (detailed) normal distributions for rendering. The MERL BRDF data set \cite{matusik2003data} contained 100 different BRDFs with real-world materials, which can provide a diverse set of surface materials for rendering shapes. The authors used a physically based ray tracer, Mitsuba \cite{jakob2010mitsuba} to render photometric stereo images. For each selected shape in these two shape data sets \cite{johnson2011shape,wiles2017silnet}, the authors of PS-FCN \cite{chen2018ps} used 1296 regularly-sampled views, randomly selected 2 of the 100 BRDFs in the MERL BRDF data set, and 64 light directions randomly sampled from the upper hemisphere space to render 64 photometric stereo images with a (cropped) spatial resolution of 128 $\times$ 128. The total number of training samples for the first method was 85212.

\subsubsection{CyclesPS data set}

The second one was proposed in CNN-PS \cite{ikehata2018cnn}, namely the CyclesPS data set.  In this data set, the authors utilized Disney’s principled BSDF data set \cite{mcauley2012practical} rather than the MERL BRDF data set \cite{matusik2003data} to provide surface reflectance. Compared to the MERL BRDF data set \cite{matusik2003data}, which had 100 measured BRDFs and thus cannot cover the tremendous real-world materials, Disney’s principled BSDF data set integrated five different BRDFs controlled by 11 parameters, which can represent a wide variety of real-world materials. Although the CyclesPS data set neglected some combinations of parameters that were unrealistic or did not strongly affect the rendering results, Disney’s principled BSDF data set can represent almost infinite surface reflectance. The number of 3D model shapes was 15, selected from the Internet under a royalty-free license. The CyclesPS data set \cite{ikehata2018cnn} used the Cycles renderer, bundled in Blender \cite{iraci2013blender}, to simulate complex light transport, and included three subsets, diffuse, specular, and metallic. Therefore, the total number of samples for training was 45. Different from PS-FCN \cite{chen2018ps}, it divided the object region in the rendered image into 5000 superpixels and used the same set of parameters at the pixels within a superpixel, \emph{i.e.}, 5000 kinds of materials in one sample. Moreover, the number of light directions was 740, which means that 740 photometric stereo images were rendered for each sample, with a spatial resolution of 256 $\times$ 256. The number of objects in the CyclesPS data set was further increased to 25 in PS-Transformer \cite{ikehata2021ps}, with the same settings.

\subsubsection{Discussion}

Compared to these two data sets, we can find that the strategy and attention are quite different. As summarized in Table \ref{data settable}, the Blobby and Sculpture data set \cite{chen2018ps} contains much more samples than the CyclePS data set\cite{ikehata2018cnn} (85212 vs. 45). However, the number of illuminated images with homogeneous reflectance is 64 in the Blobby and Sculpture data set\cite{chen2018ps}, while there are more than 700 very densely illuminated images with spatially varying materials in the CyclePS data set\cite{ikehata2018cnn}. The Blobby and Sculpture data set\cite{chen2018ps} is more suitable for all-pixel methods (see Section \ref{all}), and the CyclePS data set\cite{ikehata2018cnn} is better to be used by per-pixel methods (see Section \ref{per}). There are two reasons. First, all-pixel methods process input images in a patch-wise manner. In contrast, per-pixel methods use the observation map to learn the feature of a single pixel. Therefore, the number of samples is irrelevant as long as the spatial resolution is large enough. Second, before the introduction of the observation strategy \cite{chen2022deep}, all-pixel methods with patch-based inputs cannot handle objects with spatially varying materials, while per-pixel methods naturally avoid this problem. Therefore, previous per-pixel methods usually chose the CyclePS data set \cite{ikehata2018cnn} to optimize their models, while all-pixel methods always used the Blobby and Sculpture data set \cite{chen2018ps} (Tabulated in Table \ref{data settable}). However, the diversity of CyclePS \cite{ikehata2018cnn} is much better due to the powerful representation ability of Disney’s principled BSDF data set \cite{mcauley2012practical}, which can potentially lead to better performance of per-pixel methods using the observation maps strategy. Thus, establishing Disney’s principled BSDF \cite{mcauley2012practical} based data sets with more samples is important and urgent in future work.

\subsubsection{Settings and implementation details}

When training photometric stereo networks, the preferred optimizer was typically the Adam optimizer \cite{kingma2015adam} with default settings ($\beta_1$ = 0.9 and $\beta_2$ = 0.999) due to its excellent performance and ease of parameter tuning. However, some networks, such as HT21 \cite{honzatko2021leveraging}, opted for the RMSprop optimizer \cite{hintonneural}.

For per-pixel methods, the size of observation maps during training can influence optimization performance. In CNN-PS \cite{ikehata2018cnn}, the authors tested and found that using observation maps with the size of 32 $\times$ 32 resulted in the best performance. Consequently, subsequent per-pixel methods usually adopted this size, or a slightly larger size (\emph{e.g.}, 48 $\times$ 48 in HT21 \cite{honzatko2021leveraging}) in their training configurations. Similarly, for all-pixel methods, the size of input patches during training impacts performance and training time. Usually, the default size for input patches in all-pixel methods was set to 32 $\times$ 32 to achieve optimal results. MF-PSN \cite{liu2022deep} made a quantitative comparison of performance using different input sizes, supporting the choice of a 32 $\times$ 32 patch size as it can best balance both performance and computational costs.

Photometric stereo networks are required to handle a varying number of inputs. Consequently, choosing the number of inputs during training is also an important setting that has to be discussed. The experiments in \cite{ju2022Normattention,liu2022deep} have demonstrated that different number of inputs during training impacts the performance of photometric stereo networks. Specifically, when the number of input images used for training is close to the number used for testing, the networks will achieve better performance. In order to accommodate both sparse and dense input conditions, all-pixel methods commonly select a training input number of 32 images. In contrast, per-pixel methods often use a larger number of inputs, such as 50 to 1300, in many models that rely on observation maps. However, there are exceptions for methods specially designed for sparse inputs, such as SPLINE-Net \cite{zheng2019spline}, which utilizes 10 inputs.

\subsection{Testing data sets}

Test data sets are also needed to quantitatively evaluate the performance of different photometric stereo methods. These data sets can be divided into two categories: synthetic data sets and real data sets.

\subsubsection{Synthetic data sets}

Synthetic data sets were usually rendered with the same settings as the Blobby and Sculpture data set \cite{chen2018ps} or the CyclePS data set\cite{ikehata2018cnn}. For example, the rendered objects ``Bunny'', ``Dragon'', and ``Armadillo'' in the Stanford 3D data set\cite{curless1996volumetric} by the MERL BRDF data set \cite{matusik2003data} as well as the rendered objects ``Sphere'', ``Turlte'', ``Paperbowl'', ``Queen'', and ``Pumpkin''  by Disney’s principled BSDF data set \cite{mcauley2012practical}.

\subsubsection{Real data sets}

To effectively evaluate the robustness and performance of the presented photometric stereo methods, a better choice is to evaluate these methods on real photometric stereo images rather than synthetic images. Some data sets, such as the Gourd \& Apple data set \cite{alldrin2008photometric} and Light Stage Data Gallery \cite{goldman2009shape}, have been proposed for over a decade.  The Gourd \& Apple data set \cite{alldrin2008photometric} consisted of three objects, namely ``Apple'', ``Gourd1'', and ``Gourd2'', with 112, 102 and 98 images, respectively. The Light Stage Data Gallery \cite{goldman2009shape} consisted of six objects and 253 images were provided for each object. However, these data sets only provided calibrated light directions without ground-truth normal maps. Therefore, one can only qualitatively compare methods on these real data sets.

To quantitatively evaluate photometric stereo methods, Shi \emph{et al.} \cite{shi2019benchmark} first established a real photometric stereo data set with ground truth, namely DiLiGenT, which was the most widely used benchmark in the field of photometric stereo. This data set included ten objects with varying complexity, from simple spheres to intricate and concave geometries, and a wide range of materials, including mostly diffuse to strongly non-Lambertian surfaces with spatially varying properties. The authors illuminated and captured 96 images for each object under different lighting directions. To obtain the ground truth, the authors used a structured light-based Rexcan CS scanner, synchronized with a turn table to acquire 3D point clouds, which can calculate surface normals. Then, the shape-to-image alignment was performed to transform the 3D shape from the scanner coordinate system to the photometric stereo image coordinate system using the mutual information method in Meshlab \cite{corsini2009image}. Furthermore, the DiLiGenT benchmark \cite{shi2019benchmark} provided a test set, which is from a different viewpoint of these photoed objects (except for the object ``Ball'') using the same lighting setup. However, using a small number of objects (10) of DiLiGenT \cite{shi2019benchmark} is prone to overfitting in training deep neural networks, and the shapes scanned by a 3D scanner may have errors and blurring. 

To address these limitations, Ren \emph{et al.} \cite{ren2022diligent102} further proposed a new real-world photometric stereo data set with ground-truth normal maps, namely DiLiGenT$10^2$ because it contained 10 times larger (one hundred objects of ten shapes multiplied by ten materials) than the widely used DiLiGenT benchmark \cite{shi2019benchmark}. The authors used ten shapes to fabricate objects, from CAD models with selected materials, through a high-precise computer numerical control (CNC) machining process, rather than scanning existing objects, which greatly avoided measurement errors. For each shape, ten materials were used to make the objects, from isotropic (diffuse and specular) and anisotropic, to translucent reflectance. Recently, Wang \emph{et al.} \cite{wang2023diligent} introduced a real-world data set, DiLiGenT-$\Pi$, for detailed near-planar surfaces. This data set was specifically designed to capture objects with high-frequency detailed structures, such as coins and badges. Similar to the DiLiGenT data set \cite{shi2019benchmark}, the authors used a 3D scanner to acquire ground-truth 3D models for 30 objects in this data set. The presented training and test data sets for deep learning photometric stereo methods are summarized in Table \ref{data settable}.

\begin{table*}
\vspace*{-1mm}
    \caption{Performance on the DiLiGenT benchmark \cite{shi2019benchmark} with 96 images, measured in terms of MAE in degrees. The compared methods are ranked by the average MAE of ten objects.
    }
    \vspace*{-4mm}
    \label{benchcompare}
    \centering
    \renewcommand\arraystretch{1.2}
\begin{tabular}{ |c |c| c| c| c| c| c| c|c |c |c| c| c|}
\hline
Method& Ball& Bear& Bear-76 &Buddha &Cat& Cow& Goblet& Harvest& Pot1 & Pot2 & Reading & Avg.\\
\hline
\rowcolor{AliceBlue} Baseline \cite{Woodham1980Photometric}& 4.10&8.39& - & 14.92&8.41&25.60&18.50&30.62&8.89&14.65&19.80&15.39 \\
\rowcolor{AliceBlue} IW12 \cite{ikehata2012robust}&2.54&7.32&- &11.11&7.21&25.70&16.25&29.26&7.74&14.09&16.17&13.74 \\
\rowcolor{AliceBlue} WG10 \cite{wu2010robust}&2.06&6.50&- &10.91&6.73&25.89&15.70&30.01&7.18&13.12&15.39&13.35\\
\rowcolor{AliceBlue}HM10 \cite{higo2010consensus}&3.55&11.48&- &13.05&8.40&14.95&14.89&21.79&10.85&16.37&16.82&13.22 \\
\rowcolor{Honeydew} KS21 \cite{kaya2021uncalibrated}& 3.78&5.96&-&13.14&7.91&10.85&11.94&25.49&8.75&10.17&18.22&11.62 \\
\rowcolor{AliceBlue}IA14 \cite{ikehata2014photometric}&3.34&7.11&- &10.47&6.74&13.05&9.71&25.95&6.64&8.77&14.19&10.60 \\
\rowcolor{AliceBlue}ST14 \cite{shi2014bi}&1.74 &6.12 &-&10.60 &6.12 &13.93 &10.09 &25.44 &6.51 &8.78 &13.63 &10.30  \\
\rowcolor{MistyRose}SPLINE-Net$\dag$ \cite{zheng2019spline}&4.51&5.28&-&10.36&6.49&7.44&9.62&17.93&8.29&10.89&15.50&9.63\\
\rowcolor{Honeydew}SDPS-Net \cite{chen2019self}&2.77&6.89&-&8.97&8.06&8.48&11.91&17.43&8.14&7.50&14.90&9.51\\
\rowcolor{MistyRose}DPSN \cite{santo2017deep}&2.02&6.31&-&12.68&6.54&8.01&11.28&16.86&7.05&7.86&15.51&9.41 \\
\rowcolor{Honeydew} SK22 \cite{sarno2022neural}& 3.46&5.48&-&10.00&8.94&6.04&9.78&17.97&7.76&7.10&15.02&9.15\\
\rowcolor{MistyRose}IRPS \cite{taniai2018neural} &1.47 &5.79&-& 10.36 &5.44&6.32 &11.47 &22.59 &6.09 &7.76 & 11.03 &8.83 \\
\rowcolor{Honeydew}UPS-GCNet \cite{chen2020learned} &2.50&5.60&-&8.60&7.80&8.48&9.60&16.20&7.20&7.10&14.90&8.70\\
\rowcolor{MistyRose}LMPS \cite{li2019learning}& 2.40&5.23&-&9.89&6.11&7.98&8.61&16.18&6.54&7.48&13.68&8.41 \\
\rowcolor{MistyRose}PS-FCN \cite{chen2018ps} &2.82 &7.55 &5.02&7.91 &6.16 &7.33 &8.60 &15.85 &7.13 &7.25 &13.33 &8.39 \\
\rowcolor{Honeydew}ReDDLE-Net \cite{yang2022reddle}&2.65&6.04&-&7.28&8.76&6.80&8.42&12.28&7.82&7.99&14.03&8.21 \\
\rowcolor{MistyRose}Manifold-PSN \cite{ju2020learning}&3.05&6.31&-&7.39&6.22&7.34&8.85&15.01&7.07&7.01&12.65&8.09\\
\rowcolor{Honeydew}LERPS \cite{tiwari2022lerps}& 2.41&6.93&-&8.84&7.43&6.36&8.78&11.57&8.32&7.01&11.51&7.92\\
\rowcolor{MistyRose}Attention-PSN \cite{ju2020pay}& 2.93&4.86&-&7.75&6.14&6.86&8.42&15.44&6.92&6.97&12.90&7.92\\
\rowcolor{MistyRose}DR-PSN \cite{ju2021recovering}&2.27& 5.46&-& 7.84& 5.42& 7.01&8.49& 15.40& 7.08&7.21& 12.74& 7.90\\
\rowcolor{MistyRose}GPS-Net \cite{yao2020gps}&2.92&5.07&-&7.77&5.42&6.14&9.00&15.14&6.04&7.01&13.58&7.81\\
\rowcolor{MistyRose}JJ21 \cite{ju2021incorporating}&2.51&5.77&-&7.88&6.56&6.29&8.40&14.95&7.21&7.40&11.01& 7.80\\
\rowcolor{MistyRose}CHR-PSN \cite{ju2022learning} &2.26&6.35&-&7.15&5.97&6.05&8.32&15.32&7.04&6.76&12.52&7.77\\
\rowcolor{MistyRose}CNN-PS$\dag$ \cite{ikehata2018cnn}&2.12&8.30&4.10&8.07&4.38&7.92&7.42&14.08&5.37&6.38&12.12&7.62 \\
\rowcolor{MistyRose} SPS-Net \cite{liu2020sps} & 2.80&- &-&6.90&5.10&6.30&7.10&13.70&7.50&7.40&11.90&7.60 \\
\rowcolor{MistyRose}MT-PS-CNN \cite{cao2022learning}&2.29&5.79&-&6.85&5.87&7.48&7.88&13.71&6.92&6.89&11.94&7.56 \\
\rowcolor{AliceBlue}HS17 \cite{hui2016shape}  &1.33&5.58&-&8.48&4.88&8.23&7.57&15.81&5.16&6.41&12.08&7.55 \\
\rowcolor{MistyRose}PS-FCN (Norm.) \cite{chen2022deep}& 2.67&7.72&-&7.53&4.76&6.72&7.84&12.39&6.17&7.15&10.92&7.39\\
\rowcolor{MistyRose}MF-PSN \cite{liu2022deep}& 2.07&5.83&-&6.88&5.00&5.90&7.46&13.38&7.20&6.81&12.20&7.27\\
\rowcolor{MistyRose}HPS-Net \cite{ju2023efficient}&2.37&5.28&-&6.89&4.98&5.59&7.59&14.17&6.23&6.77&11.26 &7.11\\
\rowcolor{Honeydew}LL22b \cite{li2022self} &1.24 &3.82&-&9.28 & 4.72& 5.53&7.12&14.96&6.73&6.50&10.54 &7.05\\
\rowcolor{MistyRose}HT21$\dag$ \cite{honzatko2021leveraging}&2.49&8.96& 3.59 &7.23&4.69&4.89&6.89&12.79&5.10&4.98&11.08&6.91 \\
\rowcolor{MistyRose}PSMF-PSN \cite{Yang2023accurate} &2.54&5.99&-&7.21&5.09&5.52&7.75&11.40&6.91&6.11&10.01&6.85\\
\rowcolor{MistyRose}NormAttention-PSN \cite{ju2022Normattention} &2.93&5.48&4.80&7.12&4.65&5.99&7.49&12.28&5.96&6.42&9.93&6.83\\
\rowcolor{MistyRose}WZ20 \cite{wang2020non}&1.78&5.26&4.12 &6.09&4.66&6.33&7.22&13.34&6.46&6.45&10.05&6.76\\
\rowcolor{MistyRose}SR-PSN \cite{ju2023estimating}&2.23&5.24&-&6.75&4.63&6.12&7.07&12.61&5.88&6.44&10.35&6.73\\
\rowcolor{MistyRose} JZ23 \cite{ju2023learning} &2.26 &4.57&-&7.07&4.72&5.83&7.73&11.35&5.68&6.39&11.38&6.70\\
\rowcolor{MistyRose} IS22$\dag$ \cite{ikehata2022does}&2.30&-&3.90&7.70&4.20&5.70&7.20&13.80&5.00&5.40&10.70&6.60\\
\rowcolor{MistyRose}GR-PSN \cite{ju2023grpsn} &2.22&5.61&-&6.73&4.33&6.17&6.78&12.03&5.54&6.42&9.65&6.55\\
\rowcolor{MistyRose} LL22a \cite{li2022neural} &2.43&-&3.64&8.04&4.86& 4.72& 6.68 &14.90 &5.99& 4.97 &8.75 &6.50\\
\rowcolor{MistyRose}PX-Net$\dag$\cite{logothetis2021px} &2.03&4.13&3.57&7.61&4.39&4.69&6.90&13.10&5.08&5.10&10.26&6.33\\
\hline
\end{tabular}
\vspace*{-4mm}
\end{table*}

\section{Benchmark Evaluation Results}
The evaluation metric is based on the statistics of angular errors. For the whole normal map, the mean angular error (MAE) is calculated as follows:
\begin{equation}
\label{metrics}
\mathrm{MAE} = \frac{1}{T} \sum_{p}^{T} \mathrm{cos}^{-1} (\boldsymbol{n_p}^{\top} \boldsymbol{\tilde{n}_p}), 
\end{equation}
where $T$ is the total number of pixels on the object, excluding the pixels at background positions, and $\boldsymbol{n_p}$ and $\boldsymbol{\tilde{n}_p}$ are the ground-truth and estimated surface normal vector at the position indexed $p$. In addition to MAE, some papers also used the ratios of the number of surface normals with angular error smaller than $x^\circ$, denoted as $err_{{<x}^\circ}$ \cite{ju2020pay,ju2022Normattention}.

\begin{table*}[!htp]
\vspace*{-1mm}
\begin{center}
\caption{Performance on the DiLiGenT benchmark \cite{shi2019benchmark} with 10 images, measured in terms of MAE in degrees. The compared methods are ranked by the average MAE of ten objects. }
\vspace*{-4mm}
\label{benchsparse}
\renewcommand\arraystretch{1.2}
\begin{tabular}{ |c |c| c| c| c |c| c| c| c |c |c |c|}
\hline
Method& Ball& Bear&Buddha &Cat& Cow& Goblet& Harvest& Pot1 & Pot2 & Reading & Avg.\\
\hline
\rowcolor{AliceBlue}IA14 \cite{ikehata2014photometric}&12.94& 16.40 &20.63& 15.53 &18.08 &18.73 &32.50& 6.28& 14.31 &24.99 &19.04\\
\rowcolor{AliceBlue}Baseline \cite{Woodham1980Photometric}& 5.09&11.59&16.25& 9.66& 27.90& 19.97& 33.41 &11.32& 18.03& 19.86& 17.31\\
\rowcolor{AliceBlue}ST14 \cite{shi2014bi}&5.24& 9.39& 15.79 &9.34 &26.08& 19.71 &30.85& 9.76 &15.57& 20.08& 16.18\\
\rowcolor{AliceBlue}IW12 \cite{ikehata2012robust}&3.33 &7.62 &13.36 &8.13&25.01& 18.01 &29.37& 8.73& 14.60 &16.63 &14.48\\
\rowcolor{MistyRose}CNN-PS$\dag$ \cite{ikehata2018cnn}&9.11& 14.08 &14.58& 11.71& 14.04& 15.48 &19.56& 13.23 &14.65& 16.99 &14.34\\
\rowcolor{MistyRose}IRPS \cite{taniai2018neural}&2.12&6.92&11.41&6.58&8.87&14.99&26.55&7.14&9.61&13.70&10.79 \\
\rowcolor{MistyRose}PS-FCN \cite{chen2018ps}&4.02& 7.18& 9.79 &8.80& 10.51 &11.58& 18.70& 10.14& 9.85& 15.03& 10.51\\
\rowcolor{MistyRose}SPLINE-Net$\dag$ \cite{zheng2019spline}&
4.96 &5.99& 10.07& 7.52 &8.80 &10.43 &19.05& 8.77& 11.79& 16.13& 10.35\\
\rowcolor{MistyRose}PS-FCN (Norm.) \cite{chen2022deep}&4.38& 5.92&8.98&6.30&14.66&10.96&18.04&7.05&11.91&13.23&10.14\\
\rowcolor{MistyRose}LMPS \cite{li2019learning}&3.97 &8.73& 11.36& 6.69& 10.19& 10.46& 17.33& 7.30 &9.74 &14.37& 10.02\\
\rowcolor{MistyRose}DR-PSN \cite{ju2021recovering}&
3.83 &7.52 &9.55 &7.92 &9.83 &10.38& 17.12& 9.36& 9.16& 14.75 &9.94\\
\rowcolor{MistyRose}CHR-PSN \cite{ju2022learning}& 3.91& 7.84 &9.59& 8.10 &8.54& 10.36 &17.21 &9.65& 9.61& 14.35& 9.92\\
\rowcolor{MistyRose}MT-PS-CNN \cite{cao2022learning} &4.20&8.59&8.25&7.30&10.84&10.44&16.97&8.78&9.85&13.17&9.84\\
\rowcolor{MistyRose}JJ21 \cite{ju2021incorporating}&3.86&7.49&9.69&7.82&8.55&10.31&16.94&9.28&9.54&14.30&9.78\\
\rowcolor{MistyRose}IS22 \cite{ikehata2022does}&4.30&5.40&8.70&6.20&11.60&10.70&20.60&7.00&8.00&13.20&9.60\\
\rowcolor{MistyRose}PSMF-PSN \cite{Yang2023accurate} &3.88&5.91&8.49&6.75&11.47&9.77&16.36&8.29&11.71&12.52&9.51\\
\rowcolor{MistyRose}GPS-Net \cite{yao2020gps}& 4.33&6.34&8.87&6.81&9.34&10.79&16.92&7.50& 8.38&15.00&9.43\\
\rowcolor{MistyRose}SPS-Net \cite{yao2020gps}&4.60 &-&8.00&6.90&8.30&9.00&16.70&8.90&9.00&13.60&9.40\\
\rowcolor{MistyRose}MF-PSN \cite{liu2022deep}&2.97&4.89&7.43&5.55& 8.41&9.87&12.92&7.21&9.16&12.92&8.48\\
\rowcolor{MistyRose}PX-Net$\dag$ \cite{logothetis2021px}&2.50&4.90&9.40&6.30&7.20&9.70&16.10&7.00&7.70&13.10&8.37\\
\rowcolor{MistyRose}WJ20 \cite{wang2020non}&2.30&5.18&7.05&5.62&7.53&8.80&15.26&7.08&8.19&10.88&7.79\\
\rowcolor{MistyRose}PS-Transformer$\dag$ \cite{ikehata2021ps}&3.27&4.88&8.65&5.34&6.54&9.28&14.41&6.06&6.97&11.24&7.66\\
\hline
\end{tabular}
\end{center}
\vspace*{-5mm}
\end{table*}

In Table \ref{benchcompare}, we report the quantitative results of the above-mentioned deep learning-based calibrated (marked as red) and uncalibrated (marked as green) photometric stereo methods on the DiLiGenT benchmark data set \cite{shi2019benchmark} under all the 96 input images (dense condition). Similarly, we review the performance of these calibrated deep learning-based photometric stereo methods in the sparse condition (10 input images) tabulated in Table \ref{benchsparse}. Note that not all the methods report the results under 10 input images, and some methods only provide the sparse condition without dense input, such as PS-Transformer\cite{ikehata2021ps}. 

Besides deep learning methods, we also evaluate the performance of some representative non-learning-based calibrated algorithms (marked as blue) and compare them with deep learning-based methods. As shown in Table \ref{benchcompare}, most of the learning-based methods are represented by their networks' names. For non-learning methods and some learning-based methods without given names, we present them by the first letter of the authors' name and the published year. To ensure fairness in the evaluation, we also employ $\dag$ to denote the networks trained by CyclePS \cite{ikehata2018cnn}, which are rendered using Disney's principled BSDF data set \cite{mcauley2012practical}. Theoretically, Disney's principled BSDFs contain an extensive range of reflectance properties by integrating various BRDFs controlled by 11 parameters. Consequently, the reflectance distributions of CyclePS more closely resemble those encountered in real-world scenarios compared to the Blobby and Sculpture data set \cite{chen2018ps}, which are rendered using the MERL BRDF data set \cite{matusik2003data}. Furthermore, some recent models discarded the first 20 images of “Bear” in testing (\emph{i.e.}, tested with the remaining 76 images) because the first 20 images are photometrically inconsistent in the belly region \cite{ikehata2018cnn}. For these methods, we tabulate both the results of ``Bear'' input with 76 images and 96 images, denoted as ``Bear-76'' and ``Bear'', respectively. For a fair comparison, the average MAE of these ten objects uses the result of Bear rather than Bear-76, except for IS22 \cite{ikehata2022does} and LL22a \cite{li2022neural}, which only report the Bear-76 results. Additionally, SPS-Net \cite{liu2020sps} discards the results of Bear; therefore, we can only calculate the average MAE via the remaining nine objects. Since only parallel white lights were used in the DiLiGenT benchmark \cite{shi2019benchmark}, we can only evaluate the methods for calibrated and un-calibrated photometric stereos, ignoring methods for near-field light, general light, and color light.

\begin{figure*}
 \begin{center}
\includegraphics[width=0.9\textwidth]{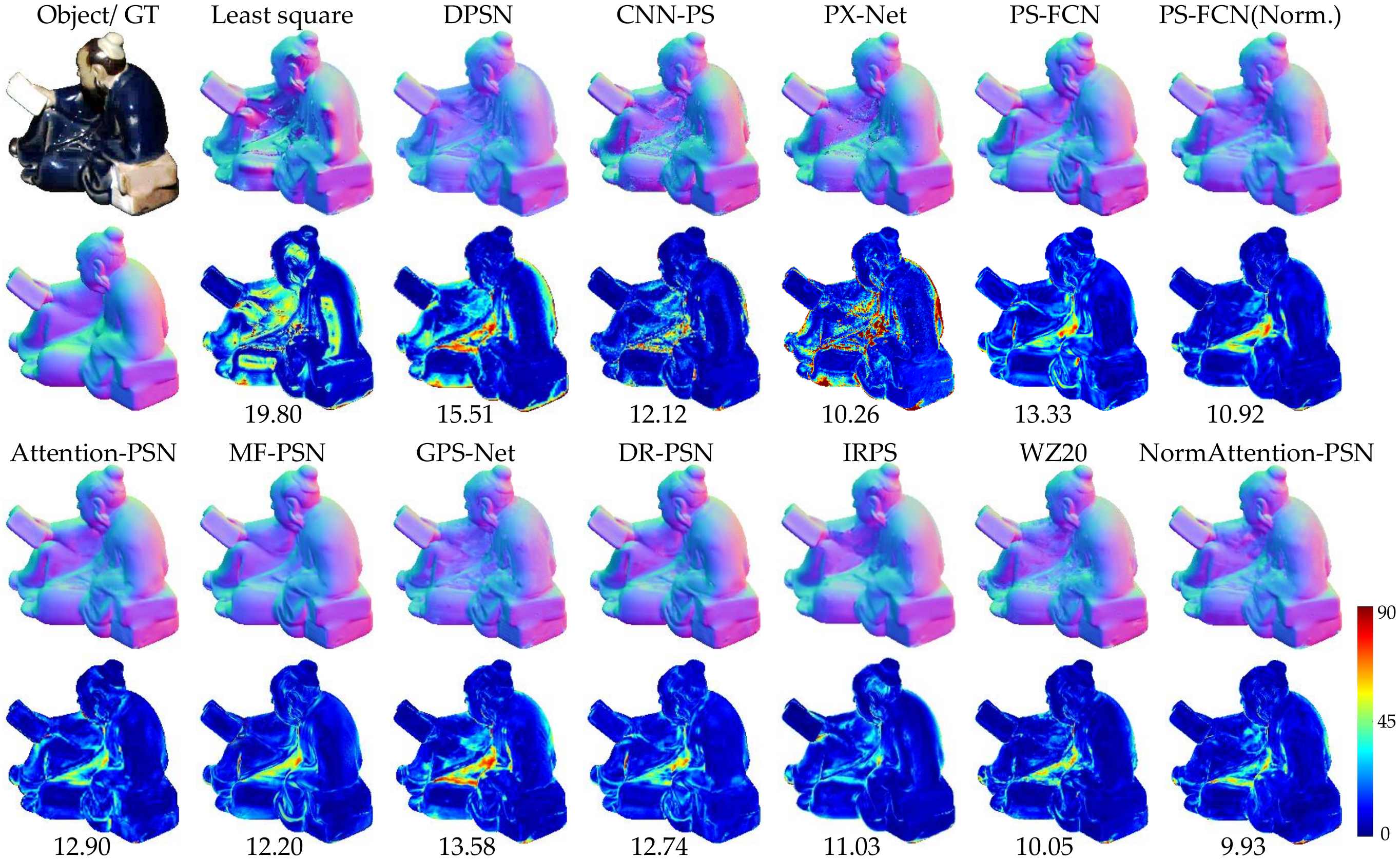}
  \end{center}
  \vspace*{-5mm}
  \caption{Quantitative results for the object ``Reading'', tested with 96 input images. The first row represents the estimated normal maps, while the second row shows the corresponding error maps, with values indicating MAE in degrees.}
  \label{figcompare}
  \vspace*{-2mm}
\end{figure*}

\begin{figure*}
 \begin{center}
  \includegraphics[width=0.9\textwidth]{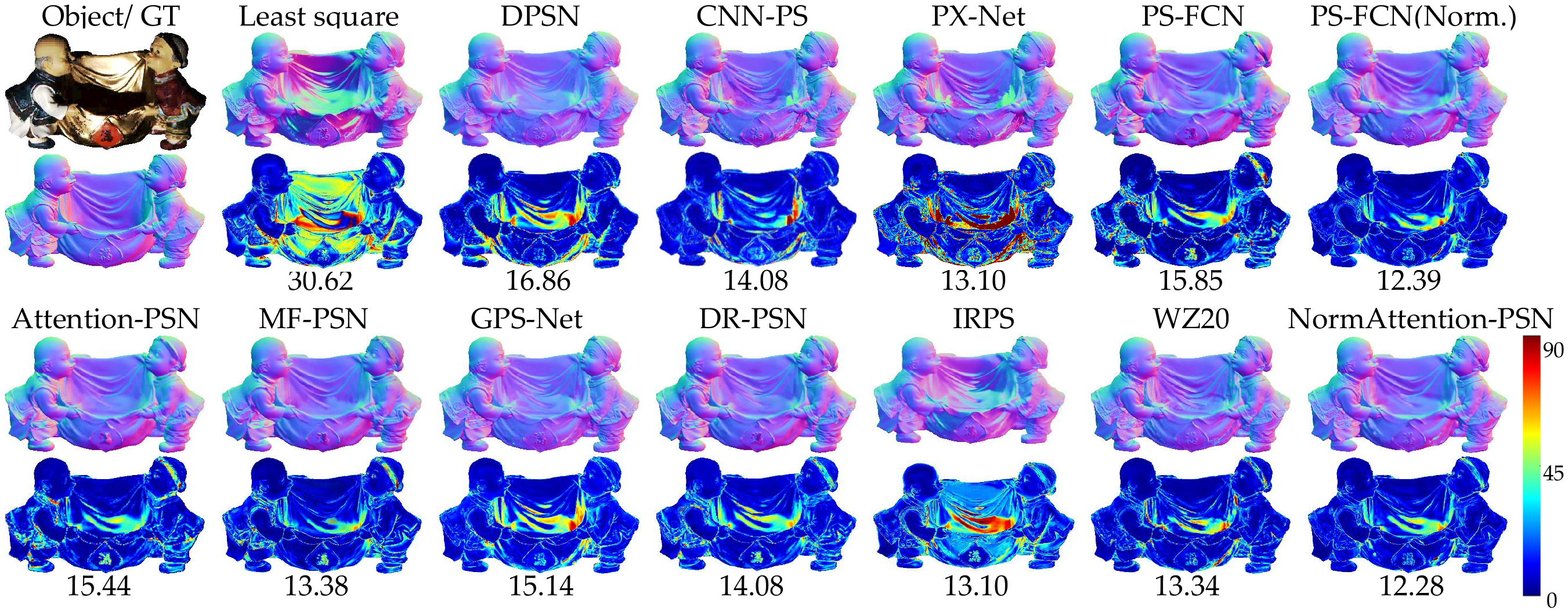}
  \end{center}
  \vspace*{-7mm}
  \caption{Quantitative results for the object ``Harvest'', tested with 96 input images. The first row represents the estimated normal maps, while the second row shows the corresponding error maps, with values indicating MAE in degrees.}
  \label{figcompare2}
  \vspace*{-3mm}
\end{figure*}

Furthermore, in Figs. \ref{figcompare} and \ref{figcompare2}, we visualize the representative deep learning-based calibrated photometric stereo methods. The visual comparisons are based on the objects  ``Reading'' and ``Harvest' in the DiLiGenT benchmark data set \cite{shi2019benchmark}. For more visualization comparisons please refer to \url{https://github.com/Kelvin-Ju/Survey-DLCPS} 

In Figs. \ref{figcompare} and \ref{figcompare2}, we evaluate the visualized reconstructed normal maps and error maps of 12 deep learning-based calibrated photometric stereo approaches according to our taxonomy, and the traditional least square method \cite{Woodham1980Photometric}. The baseline \cite{Woodham1980Photometric}, assuming Lambertian reflectance, exhibits severe errors on specular highlights. In contrast, deep learning-based methods significantly improve results in highlight regions, demonstrating the fitting ability of deep neural networks to approximate non-Lambertian surface reflectances. As the first deep network, DPSN \cite{santo2017deep} exhibits inferior results in regions with cast shadows, such as the back of the ``Reading''  and the pocket of the ``Harvest''. This limitation arises because DPSN predicts a normal vector solely based on the reflectance observations of a single pixel, neglecting information embedded in the neighborhood of a surface point. Similar issues are observed in some methods that do not consider neighboring regions \cite{li2019learning,wang2020non}. PX-Net \cite{logothetis2021px} incorporates global information into observation maps, leading to more accurate reconstruction results in shadow and highlight regions. However, the visualized normal map from \cite{logothetis2021px} exhibits sparse noises, potentially attributed to suboptimal camera noise and self-reflection settings in the generation of global effects. On the other hand, early all-pixel methods encounter errors in regions with spatially varying reflectance \cite{chen2018ps,ju2020pay,liu2022deep,ju2021recovering}, such as the edge of the hat and hair. This occurs because the convolutional network processes input images in a patch-wise manner, where steep color changes impact the entire patch, such as the hat of the ``Reading'' and the cloth of the ``Harvest''. This problem is eventually addressed by the normalization operation in PS-FCN (Norm.) \cite{chen2022deep} and the double-gate normalization in NormAttention-PSN \cite{ju2022Normattention}, which can better handle color-changed surfaces.

Furthermore, as displayed in Tables \ref{benchcompare} and \ref{benchsparse}, the results based on deep learning-based photometric methods generally achieve better performance, as compared with non-learning methods, especially in objects with complex structure and strong non-Lambertian reflectance (``Harvest'', ``Reading''). This illustrates the capability and generalization of deep learning techniques. However, it can be seen that most deep learning models achieve ordinary performance on very simple objects with almost diffuse reflectance, such as ``Ball''. We believe that this may result from overfitting in ``complex'' network structures and ``difficult'' BRDF training data sets \cite{matusik2003data,mcauley2012practical} that pay more attention to non-Lambertian materials \cite{zheng2019spline}.

\section{Future Trends}
In this section, we point out some promising trends for future development, based on the discussion in the above sections. First, we focus on the problem of calibrated photometric stereo. Then, we raise the perspective of the entire photometric stereo community. 

As discussed in Section \ref{inputclass}, we compare the unique characteristics of per-pixel and all-pixel methods. These methods can be further explored and better combined. For per-pixel methods, we believe that some further developments can be found in observation maps \cite{ikehata2018cnn}, \emph{e.g.}, how to optimize unstructured light vectors via a graph-based network \cite{wu2019simplifying} in the ``observation map'', how to embed the information from adjacent surface points in a per-pixel manner. For all-pixel methods, we believe that the fusion of inter-images (inter-patches) still needs to be improved. Existing methods applied max-pooling \cite{chen2018ps,ju2022Normattention,liu2022deep} or manifold learning \cite{ju2020learning} to aggregate a flexible number of input images. However, these methods are underutilized for fusing features or suffer from cumbersome training pipelines. Therefore, a better fusion strategy should be proposed, which can leverage the self-attention mechanism \cite{ashish2017attention} to learn the weights of input features. Of course, a more far-sighted research direction is how to efficiently combine per-pixel and all-pixel methods, which has been initially discussed in recent combined works \cite{honzatko2021leveraging,cao2022learning}, and can be further explored by mutually combining with more physical cues. Furthermore, we argue that deep learning photometric stereo models can be further improved by excavating prior knowledge \cite{ju2021incorporating,wang2020non} and supervisions \cite{ju2020dual,tiwari2022lerps}. 

In fact, many deep learning-based photometric stereo methods discussed above are calibrated photometric stereo algorithms, which assume stringent requirements, such as accurate directions of incident illuminations, directional illuminations, and standard darkrooms, \emph{etc}. In practical applications, many assumptions are not satisfied. When reviewing the realistic environment, we naturally expect a general or universal model that can handle un-calibrated light \cite{chen2019self,sarno2022neural}, the colored light \cite{antensteiner2019single,ju2020dual}, the near-field light \cite{santo2020deep2,logothetis2022cnn}, the general light \cite{hold2019single}, and even a perspective projection camera simultaneously. Recently, an inspiring work UniPS\cite{ikehata2022universal} first dropped the physical lighting models and extracted a generic lighting representation in image interaction.  This enables UniPS to accommodate various lighting scenarios, including parallel lighting, spatially varying lighting, near-field lighting, and outdoor wild lighting. Additionally, Ikehata introduced SDM-UniPS \cite{ikehata2023scalable}, designed for high-resolution input images and considering non-local interactions among surface points. SDM-UniPS \cite{ikehata2023scalable} achieves scalable, detailed, and mask-free photometric stereo reconstruction under a universal light environment. However, these methods may face limitations when handling photometric stereo images with minor variations in lighting. This limitation arises from their reliance on the interaction mechanism for learning global lighting context rather than extracting features from each individual input. In this direction, we believe that more work can explore more effective extraction ways of universal lighting.

Furthermore, recent advancements in neural rendering technologies, \emph{i.e.}, Neural Radiance Fields (NeRF) \cite{mildenhall2020nerf}, have demonstrated great potential in photometric stereo when integrated with multi-view reconstruction. \cite{kaya2022neural,yang2022ps} proposed NeRF-based multi-view photometric stereo methods, which first estimate per-view surface normal maps and then blend them with a multi-view neural radiance field representation to reconstruct the object’s surface geometry. Multi-view photometric stereo methods \cite{kaya2022neural,yang2022ps} can offer a comprehensive 3D shape perception, while almost all single-view photometric stereo methods fail to recover the invisible parts ($\mathrm{S^{3}}$-NeRF \cite{yang2022s} can learn a neural scene representation to recover the invisible 3D parts via the single-view photometric stereo images). Notably, these NeRF-based multi-view photometric stereo techniques can avoid noticeable accumulated errors compared to traditional multi-view photometric stereo methods, which typically involve multiple disjoint and complex stages. However, existing NeRF-based photometric stereo methods still have limitations and could be explored as future trends. Firstly, NeRF-based photometric stereo methods impose a substantial computational burden and require lengthy retraining for new objects. Secondly, these methods take multi-view multi-light photometric stereo images as input, which involves fixing the camera at each viewpoint while varying the light directions. We argue that more works can explore the neural rendering techniques based on multi-view single-light, \emph{i.e.}, light can be associated with the moving camera, potentially enhancing usability in real-world applications.

\section{Conclusion}
In this paper, we conducted a systematic review of deep learning-based photometric stereo methods. According to our taxonomy focusing on calibrated deep learning-based photometric stereo methods, we have summarized and discussed the strengths and weaknesses of these models by categorizing them by input processing, supervision, and network architecture. We also introduced the used training data sets and test benchmarks in the field of photometric stereo. Then, more than thirty calibrated and uncalibrated deep learning models for photometric stereo were evaluated on the widely used benchmark. Compared with traditional non-learning methods, deep learning-based photometric stereo models are superior in estimating surface normals. Finally, we pointed out the future trends in the field of photometric stereo. We hope that this survey will help researchers orient themselves to develop in this growing field, as well as highlight opportunities for future research.

\ifCLASSOPTIONcaptionsoff
  \newpage
\fi

\bibliographystyle{IEEEbib}
\bibliography{bibinfo}

\end{document}